%% file: cas-dc-template.tex
\documentclass[a4paper,fleqn]{cas-dc}

\usepackage[numbers]{natbib}
\usepackage{bm}

\def\tsc#1{\csdef{#1}{\textsc{\lowercase{#1}}\xspace}}
\tsc{WGM}
\tsc{QE}
\tsc{EP}
\tsc{PMS}
\tsc{BEC}
\tsc{DE}

\begin{document}
\let\WriteBookmarks\relax
\def\floatpagepagefraction{1}
\def\textpagefraction{.001}

\shorttitle{Benchmarking NeRF for Autonomous Robots}

\shortauthors{Ming et~al.}

\title [mode = title]{Benchmarking Neural Radiance Fields for Autonomous Robots: An Overview}                      



%
\author[hdu]{Yuhang Ming}\fnmark[1]\cormark[1]
\author[cardc]{Xingrui Yang}\fnmark[1]
\author[sit]{Weihan Wang}\fnmark[1]
\author[indiana]{Zheng Chen}\fnmark[1]
\author[ccny]{Jinglun Feng}\fnmark[1]
\author[bristol]{Yifan Xing}
\author[zju]{Guofeng Zhang}

\fntext[fn1]{Equal Contribution}
\cortext[cor1]{Corresponidng Author: yuhang.ming@hdu.edu.cn}

\affiliation[hdu]{organization={School of Computer Science, Hangzhou Dianzi University},
            city={Hangzhou},
            postcode={310018}, 
            country={China}}
\affiliation[cardc]{organization={High-speed Aerodynamics Institute, CARDC},
            city={Mianyang},
            postcode={621000}, 
            country={China}}
\affiliation[sit]{organization={Stevens Institute of Technology},
            city={Hoboken},
            postcode={07030},
            state={NJ},
            country={USA}}
\affiliation[indiana]{organization={Luddy School of Informatics, Computing, and Engineering, Indiana University},
            city={Bloomington},
            postcode={47405},
            state={IN},
            country={USA}}
\affiliation[ccny]{organization={City College of New York},
            city={New York},
            postcode={10031},
            state={NY},
            country={USA}}
\affiliation[bristol]{organization={School of Computer Science, University of Bristol},
            city={Bristol},
            postcode={BS8 1UB},
            country={UK}}
\affiliation[zju]{organization={State Key Lab of CAD\&CG, Zhejiang University},
            city={Hangzhou},
            postcode={310058},
            country={China}}

\begin{abstract}
Neural Radiance Fields (NeRF) have emerged as a powerful paradigm for 3D scene representation, offering high-fidelity renderings and reconstructions from a set of sparse and unstructured sensor data. In the context of autonomous robotics, where perception and understanding of the environment are pivotal, NeRF holds immense promise for improving performance.
In this paper, we present a comprehensive survey and analysis of the state-of-the-art techniques for utilizing NeRF to enhance the capabilities of autonomous robots. 
We especially focus on the perception, localization and navigation, and decision-making modules of autonomous robots and delve into tasks crucial for autonomous operation, including 3D reconstruction, segmentation, pose estimation, simultaneous localization and mapping (SLAM), navigation and planning, and interaction. Our survey meticulously benchmarks existing NeRF-based methods, providing insights into their strengths and limitations.
Moreover, we explore promising avenues for future research and development in this domain. Notably, we discuss the integration of advanced techniques such as 3D Gaussian splatting (3DGS), large language models (LLM), and generative AIs, envisioning enhanced reconstruction efficiency, scene understanding, decision-making capabilities.
This survey serves as a roadmap for researchers seeking to leverage NeRFs to empower autonomous robots, paving the way for innovative solutions that can navigate and interact seamlessly in complex environments.
\end{abstract}




\maketitle

\section{Introduction}
\label{sec::intro}
\input{sections/1_Introduction}

\section{Background}
\label{sec::bg}

\input{sections/2_Background}

\section{3D Reconstruction}
\label{sec::recon}

\begin{figure}[t]
    \centering
    \includegraphics[width=\linewidth]{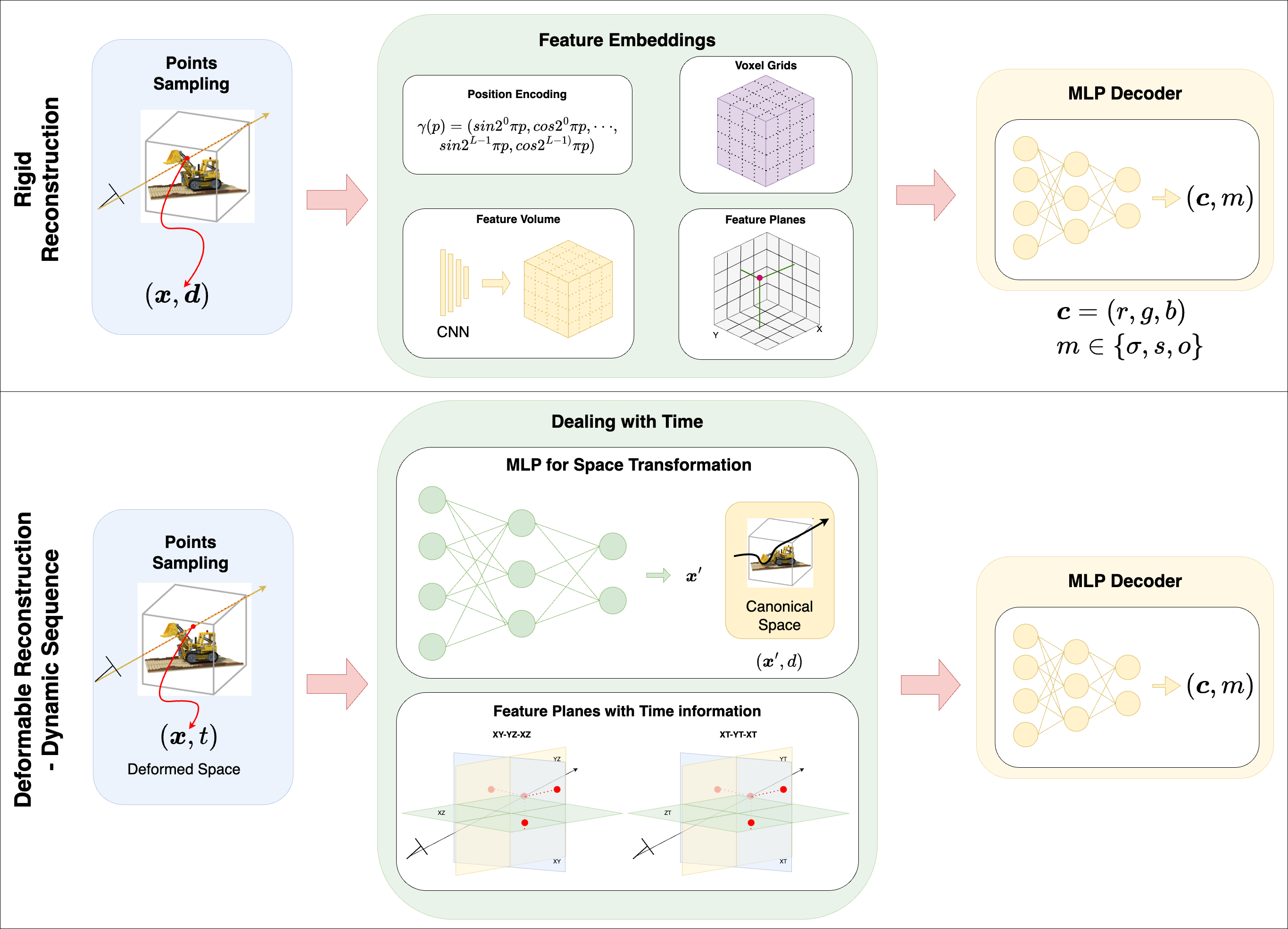}
    \caption{Overview of NeRF-based 3D reconstruction pipeline.}
    \label{fig:recon}
\vspace{-4ex}
\end{figure}

In autonomous robotics, 3D reconstruction serves as a cornerstone ability as it offers robots a comprehensive spatial understanding of the physical world they inhabit. In particular, by creating a precise 3D map of the environment, robots gain invaluable insights into the intricate details, spatial relationships, and structural layouts. This capability fuels their capacity for effective localization, navigation, manipulation, and interaction tasks.
In this section, we provide a detailed review of using neural radiance fields for 3D reconstruction. Depending on the nature of the target scene, we classify the existing methods into \textit{rigid reconstruction} and \textit{deformable reconstruction} with an overview of the pipeline shown in Figure~\ref{fig:recon}.

\subsection{Rigid Reconstruction}
\label{subsec::rigid}

\input{sections/4.1_rigid_recon}

\subsection{Deformable Reconstruction}
\label{subsec::deform}
\input{sections/4.2_deformable_recon_v3}

\subsection{Datasets and Evaluations}
\label{subsec::recon_eval}
\input{sections/4.3_recon_eval}

\section{Segmentation}
\label{sec::segmentation}
\input{sections/7_segmentation}

\section{Pose Estimation}
\label{sec::pose}
\input{sections/6_pose_estimation}

\section{SLAM}
\label{sec::slam}
\input{sections/8_SLAM}

\section{Planning and Navigation}
\label{sec::plan-navi}
\input{sections/9_plan_navi}

\section{Interaction}
\label{sec::interac}

\input{sections/10_interaction}

\section{Future Directions}
\label{sec::future}
\input{sections/11_future_works}

\section{Conclusion}
\label{sec::conclusion}
\input{sections/12_conclusion}



\bibliographystyle{model1-num-names}

\bibliography{cas-refs}





\end{document}

%% file: sections/1_Introduction.tex
In recent years, the intersection of artificial intelligence, especially deep learning techniques, and autonomous robotics has catalyzed significant advancements, revolutionizing the way robots perceive and interact with their surrounding environments. Among the myriad techniques propelling this progress, neural radiance fields (NeRF) stand out as a particularly promising innovation. 

Introduced by Mildenhall \textit{et al.}~\cite{nerf-c}, NeRF represents a groundbreaking approach in the field of novel view synthesis. 
The core principle of NeRF involves modeling the 3D scene with neural networks.
It utilizes multi-layer perceptrons (MLPs) to map 5D inputs, representing the spatial positions (3D) and viewing directions (2D), to the appearance and geometry of the scene, which is originally represented by volume density and RGB color. This mapping, trained on a sparse set of images, enables photo-realistic synthesis of novel views
of complex scenes with unprecedented detail. 
More recently, the transformative potential of NeRF has become increasingly apparent across various related fields of research.
Especially, in autonomous robotics, NeRF is becoming pivotal as it largely enhances robotic perception, thereby enabling more sophisticated and nuanced interactions between robots and their environments.

Traditional robotic perception methods heavily rely on input sensor data, such as depth maps or point clouds, to reconstruct the observed scenes and inform decision-making processes. While effective in controlled environments, these approaches suffer severe performance degradation when the input sensor data is limited or incomplete, or when the robots operate in complex, dynamic environments with serious variability and uncertainty. 
NeRF, however, offers a novel paradigm by 
representing 3D scenes with MLPs. Such a design is capable of reconstructing high-fidelity 3D models from limited data,
thereby circumventing the limitations of traditional sensor data-dependent methods.

It is worth noting that autonomous robots are complicated systems. Various robot architectures have been proposed during the decades of developments~\cite{arkin1998behavior, kortenkamp1998artificial, oreback2004component}. For simplification, we summarize six core modules for any autonomous robots, namely \textit{perception}, \textit{localization and navigation}, \textit{decision-making}, \textit{execution}, \textit{communication}, and \textit{state-monitoring and fault-handling}. Among these modules, plenty of works have focused on the applications of NeRF in the first three modules and demonstrated compelling improvements over traditional approaches. 


Starting with robotic perception, one of the primary areas is rigid body reconstruction, where NeRF helps in creating detailed 3D models of articulated objects or large-scale scenes. This capability is crucial for robots tasked with precise manipulation, assembly, or inspection. In deformable reconstruction, NeRF extends its utility to deformable objects and dynamic scenes for wider applications and better robustness.
Additionally, NeRF contributes to the segmentation of complex scenes, helping robots recognize and differentiate between multiple objects and surfaces at a granular level.

Moving on to localization and navigation, NeRF also assists in pose estimation and 
simultaneous localization and mapping (SLAM). 
In particular, in pose estimation, NeRF is a perfect solution to solve the insufficient data issue when training the pose regression networks.
Leveraging its ability to represent the 3D scene, NeRF contributes to pose estimation and SLAM tasks by tracking camera pose and, optionally, optimizing 3D scenes through iterative updates using gradient-based optimization via differentiably rendering RGB and depth information if provided.
Then, NeRF facilitates advanced navigation strategies by providing detailed environmental models that help robots anticipate obstacles and plan paths efficiently. 

Finally, in decision-making, NeRF aids in active SLAM, allowing robots to explore and map unfamiliar environments in real-time. This capability is vital for autonomous vehicles and drones that operate in dynamic or unstructured settings. Furthermore, NeRF as a rich scene representation allows robots to better interact with environments, facilitating learning of common manipulation tasks such as object grasping, policy learning, etc.

In this paper, we delve into the integration of NeRF within these three modules of autonomous robots, exploring the approaches, benchmarking the performance, and discussing challenges. In addition, we will also
investigate future work at the intersection of NeRF and other cutting-edge technologies. The integration of NeRF with techniques like 3D Gaussian splatting (3DGS), large language models (LLMs), diffusion models, and generative adversarial networks (GANs) opens new possibilities for further enhancing the capabilities of robots. 


\begin{figure}[t]
    \centering
    \includegraphics[width=\linewidth]{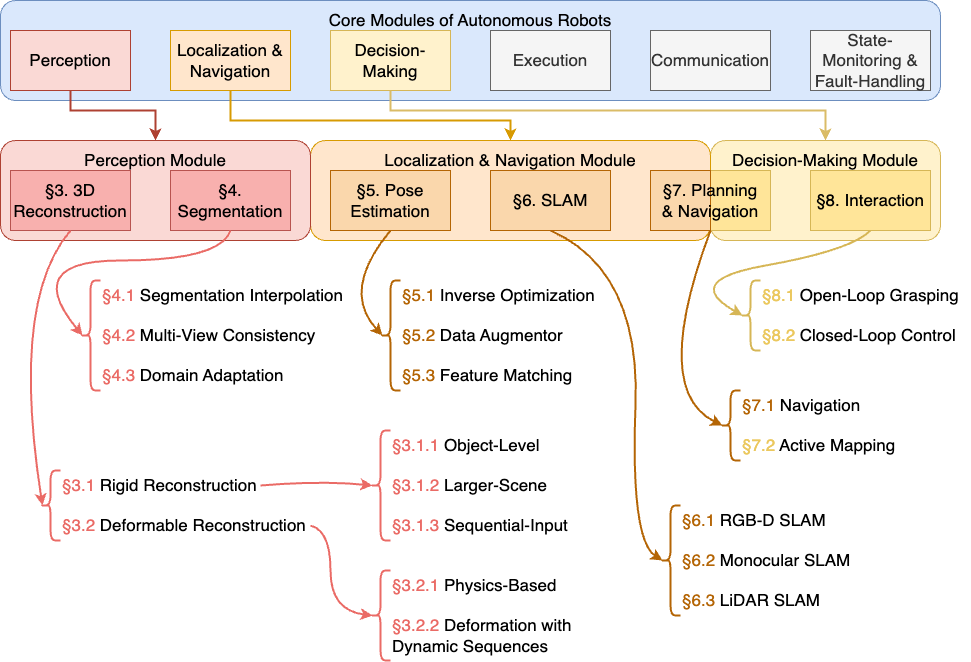}
    \caption{The structure of NeRF for Autonomous Robots: 3D Reconstruction ($\S$\ref{sec::recon}), Segmentation ($\S$\ref{sec::segmentation}), Pose Estimation ($\S$\ref{sec::pose}), SLAM ($\S$\ref{sec::slam}), Planning and Navigation ($\S$\ref{sec::plan-navi}), and Interaction ($\S$\ref{sec::interac}).}
    \label{fig:taxonomy}
\end{figure}

In the following sections, we first review preliminaries of NeRF and related existing surveys in Background ($\S$\ref{sec::bg}). Then, we thoroughly review the utilization of NeRF in 6 robotic tasks, 3D Reconstruction ($\S$\ref{sec::recon}), Segmentation ($\S$\ref{sec::segmentation}), Pose Estimation ($\S$\ref{sec::pose}), SLAM ($\S$\ref{sec::slam}), Planning and Navigation ($\S$\ref{sec::plan-navi}), and Interaction ($\S$\ref{sec::interac}). Appending to these sections, we also provide a discussion on going beyond with NeRF for robotic tasks in Future Directions ($\S$\ref{sec::future}). A taxonomy of this survey is shown in Figure~\ref{fig:taxonomy}.



%% file: sections/2_Background.tex
\begin{figure}[t]
    \centering
    \includegraphics[width=\linewidth]{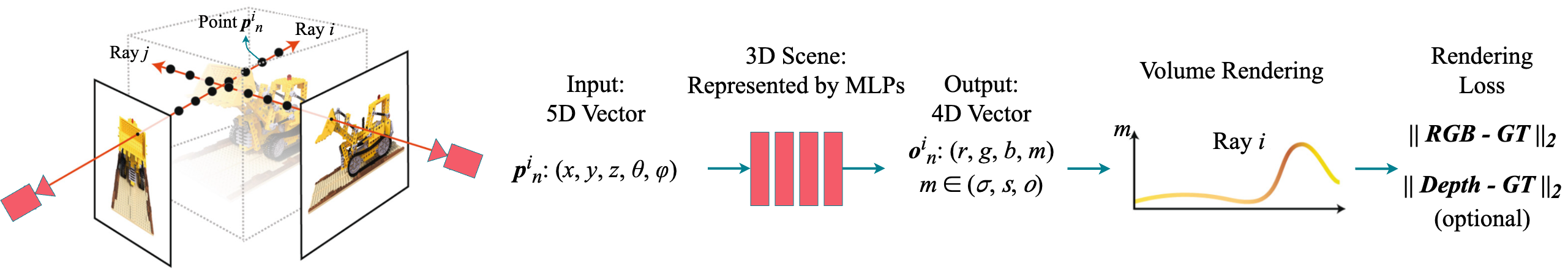}
    \caption{Overview of NeRF. Figure modified from \cite{nerf-c}}
    \label{fig:nerf-pipeline}
\vspace{-4ex}
\end{figure}

With the overview of NeRF illustrated in Figure~\ref{fig:nerf-pipeline}, we first review the key ideas and components of NeRF and its follow-up works in this section. Then, we summarize the existing NeRF-related surveys and provide key contributions of our survey. 

\subsection{NeRF Review}

\textbf{Scene representation:}
At its core, NeRF encodes 3D scenes in the weights of neural networks. 
In other words, it uses MLPs to form the mapping from spatial location to the implicit surface representation of targeting objects or scenes. In particular, the original NeRF~\cite{nerf-c, nerf-j} opts for a 5D representation $(\bm{x}, \bm{d})$ for each input, where $\bm{x}=(x, y, z)^T$ and $\bm{d}=(\theta, \phi)^T$ are the 3D position coordinates and 2D viewing direction respectively. Regarding the output of the MLP, the original NeRF favors the volume density $\sigma_i$ as the implicit surface representation. However, other implicit representations like (truncated) signed distance function (SDF) $s_i$ or occupancy $o_i$ are also widely adopted in the follow-up works~\cite{neus, azinovic2022cvpr, NICESLAM}.
With the additional color output $\bm{c} = (r, g, b)^T$, the final 4D output of the MLP is $(\bm{c}, m)$ where $m$ is chosen from $[\sigma, s, o]$. Assuming the weights of the MLP are $\Theta$, then the scene representation of NeRF can be summarized as:
\begin{equation}
    \bm{F}_\Theta(\bm{x}, \bm{d}) = (\bm{c}, m)
\end{equation}

\textbf{Positional encoding:}
As noted in the NeRF paper~\cite{nerf-c}, when directly operating on the 5D input $(\bm{x}, \bm{d})$, the network tends to overlook the high-frequency components in the scene, resulting in over-smoothed output views. To solve this issue, various positional encoding methods have been proposed to map the 5D input to a higher-dimensional feature space. The original NeRF, heuristically, chooses the sinusoidal functions:
\begin{equation}
    \gamma(\bm{p}) = (\text{sin}(2^0\pi\bm{p}), \text{cos}(2^0\pi\bm{p}), \dots, \text{sin}(2^{L-1}\pi\bm{p}), \text{cos}(2^{L-1}\pi\bm{p}))
\end{equation}
where $\bm{p}$ is chosen from $[\bm{x}, \bm{d}]$. Then, in a follow-up work~\cite{tancik2020fourfeat}, Fourier mapping with the following form is also explored:
\begin{equation}
    \gamma(\bm{p}) = (\text{sin}(2\pi\bm{B}\bm{p}), \text{cos}(2\pi\bm{B}\bm{p}))
\end{equation}
where every entry in $\bm{B}\in\mathbb{R}^{L\times d}$ is sampled from a zero-mean Gaussian whose covariance is a task-related hyperparameter.

\textbf{Volume rendering:}
To recover explicit color and geometry for any camera ray, classical volume rendering~\cite{volrender} is employed. Assume the camera ray is $\bm{r}(t)=\bm{o}+t\bm{d}$ whose near and far bounds are $t_n$ and $t_f$, where $o$ is the camera center. With the scene implicitly represented by volume density, as in the original NeRF~\cite{nerf-c}, $N$ samples $t_i$ are first uniformly sampled in between $t_n$ and $t_f$. Then, the color of the camera ray $\hat{C}(\bm{r})$ is estimated with:
\begin{equation}
    \hat{C}(\bm{r}) = \sum_{i=1}^NT_i(1-\text{exp}(-\sigma_i\delta_i))\bm{c}_i, \;\;\; \text{where} \;\;\; T_i=\text{exp}(-\sum_{j=1}^{i-1}\sigma_i\delta_i)
\end{equation}
where $\delta_i = t_{i+1} - t_i$ is the distance between two adjacent samples.

When using the signed distance function, the above equation, following~\cite{azinovic2022cvpr}, becomes:
\begin{equation}
    \hat{C}(\bm{r}) = \frac{1}{\sum_{i=1}^{N} w_i}\sum_{i=1}^N w_i\bm{c}_i, \;\;\; \text{where} \;\;\; w_i = \sigma(\frac{s_i}{tr})\cdot\sigma(-\frac{s_i}{tr})
\end{equation}
where $\sigma(\cdot)$ is the sigmoid function, $tr$ is the truncation distance

Finally, with occupancy, as in~\cite{NICESLAM}, the color can be estimated as:
\begin{equation}
    \hat{C}(\bm{r}) = \sum_{i=1}^N w_i\bm{c}_i, \;\;\; \text{where} \;\;\; w_i = o_i \prod_{j=1}^{i-1}(1-o_j)
\end{equation}

\subsection{Related survey}
Several survey papers have contributed to the understanding of topics closely related, but still essentially different to NeRF in robotics. 
To start with, many existing surveys focused on various components of the NeRF model and reviewed many papers regarding novel view synthesis performance. For instance, Rabby \textit{et al.} paid more attention to the scalability of NeRF models, covering articulated objects and city-scale environments~\cite{rabby2024beyondpixels}, 
Zhu \textit{et al.} focused on progress in volumetric rendering and factorizable embedded space~\cite{zhu2023SIP}, and Gao \textit{et al.} slightly extended the scope to the combination with GANs and diffusion models~\cite{gao2023nerf3dv}. Despite that, the latter two surveys provided a discussion on robotics-related applications like pose estimation, SLAM, semantic segmentation, and 3D reconstruction, neither the breadth nor width of the discussion is sufficient from a robotic perspective. 

Moving on to downstream applications, Nguyen \textit{et al.} targeted the visual scene understanding with NeRF, covering semantic segmentation, object pose estimation, and combination with language models, \textit{etc.}~\cite{nguyen2024semanticallyaware}. Although there are overlaps between this survey and ours, we present a broader scope with not only semantic perception but also geometric perception and decision-making in robotics. On the contrary, Remondino \textit{et al.} addressed the NeRF-based reconstruction by evaluating the strengths and weaknesses of using NeRF for 3D reconstruction~\cite{Remondio2023remotesensing}. However, this paper missed a big chunk of recent advances. Diving into more application-driven usages, Arshad \textit{et al.}, Croce \textit{et al.}, and Molaei \textit{et al.} surveyed the advances in using NeRF for 3D plant geometry reconstruction~\cite{arshad2024evaluating}, digital heritage reconstruction~\cite{Croce2024remotesensing}, and medical image reconstruction~\cite{molaei2023medicalnerf}, illustrating the necessity of reviewing NeRF in specific tasks. 

More recently, 3DGS, which is a significant departure from NeRF, has attracted a lot of attention. Following this trend, Fei \textit{et al.}~\cite{fei2024arxiv3d}, Chen \textit{et al.}~\cite{chen2024survey} and Tosi \textit{et al.}~\cite{tosi2024nerfs-gs-slam} posted surveys covering the booming developments of 3DGS in robotics tasks like SLAM, reconstruction, dynamic scene modeling. Although we also believe 3DGS is a promising direction for future research, we think there lack a thorough review and benchmark of NeRF for autonomous robots.








While these surveys offer valuable insights from computer vision perspectives, there remains a notable gap in the literature concerning the specific application of NeRF in autonomous robotic tasks. 
Therefore, this survey paper concentrates on autonomous robotics and aims to bridge the gap by providing an in-depth analysis of the role of NeRF in various robotic tasks. 
Especially, this survey targets perception, localization and navigation, and decision-making, which we believe NeRF can provide significant advantages over traditional approaches.

Concurrent to our work, He \textit{et al.} and Wang \textit{et al.} also surveyed the applications of NeRF in autonomous driving~\cite{he2024nerfad} and robotics~\cite{wang2024nerfrobotics}, and these two are the closest surveys to our work. However, there are some significant differences. Although the scopes are overlapped to some extent, He \textit{et al.} primarily focus on the outdoor scenes, ignoring a lot of work for indoor scenarios. On the other hand, Wang \textit{et al.} emphasize discussing the approaches of each method, while we not only categorize the existing approaches but also summarize the reported performance of each method for benchmarking. Additionally, we provide a more in-depth review of the existing works. For example, Wang \textit{et al.} only covered a small number of SLAM (up to NICER-SLAM~\cite{zhu2023nicer}) in rigid reconstruction, while a large number of works have emerged after NICER-SLAM, let alone there are plethora of works on 3D reconstruction itself. Thus, in our review, we separate 3D reconstruction from SLAM and provide a more thorough review of existing methods.

%% file: sections/4.1_rigid_recon.tex
Rigid reconstruction focuses on reconstructing 3D models of objects or environments with static geometry. The majority of NeRF-based rigid reconstruction methods follow the original NeRF model~\cite{nerf-c, nerf-j} and use batched views of the scene as input. Depending on the size of the scene, we first review the methods for \textit{object-level reconstruction} and then the ones for \textit{larger-scene reconstruction}. Following this we provide additional reviews on rigid reconstructions with sequential input in \textit{sequential-input reconstruction}. Finally, we summarize the datasets, and metrics used for evaluation in \textit{Datasets and Evaluations}, along with results from the reviewed papers.



\subsubsection{Object-Level}

\textbf{UNISURF}~\cite{unisurf} and \textbf{NeuS}~\cite{neus} are the first works to combine object-level neural surface reconstruction with NeRF~\cite{nerf-c, nerf-j}. They alleviate the requirements of the per-pixel object masks of the existing neural surface reconstruction methods like IDR~\cite{idr} and DVR~\cite{dvr}, and improve the surface reconstruction quality of NeRF by using a continuous occupancy field and an SDF respectively. Building on these, improvements are made in the following directions: \textit{better reconstruction quality}, \textit{sparser input views}, and \textit{reflective object surface}.

To create a better geometry representation, \textbf{VolSDF}~\cite{volsdf} particularly favors volume density and models it as a function of transformed SDF. 
On the other hand, \textbf{Geo-NeuS}~\cite{geoneus} pinpoints the gap between volume rendering integral and point-based SDF modeling. It proposes to bridge the gap with explicit SDF supervision from point clouds on the SDF network and additional geometry-consistent photometric supervision from multi-view stereo. 
Additionally, \textbf{PSDF}~\cite{psdf} takes advantage of the external geometric priors, \textit{i.e.} normals, extracted by pre-trained multi-view stereo models, creating extra supervision. It also introduces feature-level multi-view consistency based on the extracted normals.  
Focusing on the high-frequency details of the objects, \textbf{HF-NeuS}~\cite{hfneus} observed that jointly encoding high-frequency and low-frequency components in a single SDF often leads to unstable optimization. Therefore, it decomposes the SDF surface representation into base and displacement functions with a coarse-to-fine learning strategy to concentrate on regions near the surface. 
Also addresses the unstable optimization caused by high-frequency components, \textbf{Jiang \textit{et al.}}~\cite{jiang2023iccv} discretized the continuous coordinates with the nearest interpolation when training NeRF. Such a simple technique can be applied to existing methods with improved performance.
Nevertheless, rather than memorizing all the high-frequency details present in the scene with the radiance network, \textbf{NeuralWarp}~\cite{neuralwarp} proposed to directly warp images onto each other views relying only on the geometry network. 
Similarly, \textbf{LoD-NeuS}~\cite{lodneus} introduced a multi-scale tri-plane-based scene representation to share the burden of handling the complex scene geometry. In addition, cone discrete sampling and multi-convolved featurization are proposed to aggregate features within a cone frustum.
Taking a step beyond reconstruction quality, \textbf{Voxurf}~\cite{voxurf} concentrates on the training efficiency. It employs a hybrid scene representation with two separate voxel grids for SDF and features, and a dual decoder network, achieving 20x training speedup.

In order to get reconstruction with sparser input views, \textbf{SparseNeuS}~\cite{sparseneus} proposed to learn generalizable priors from image features and cascade geometry reasoning scheme for generic surface prediction. Specifically, the coarse volume encodes global features for fundamental geometries. Then, fine volumes guided by the coarse volume are constructed to refine the geometry. 
Also using image features as priors, \textbf{VolRecon}~\cite{volrecon} creates feature volumes similar to NeuralRecon~\cite{NeuralRecon}. The feature volumes across different views are aggregated with a ViewTransformer and features along a ray are obtained with a RayTransformer, which are later decoded into the signed ray distance function (SRDF) with an MLP for explicit surface reconstruction.
Taking it to multiple resolutions, \textbf{GenS}~\cite{gens} creates multi-scale feature volumes and introduces the multi-scale feature-metric consistency to constrain the geometry, achieving better performance than photometric consistency.

From another perspective, lots of efforts have also been made to improve the reconstruction of reflective objects. Build on factorized tensor components, TensoRF~\cite{tensorf} and neural incident light field, NeILF~\cite{neilf}, \textbf{TensoIR}~\cite{tensoir} and \textbf{NeILF++}~\cite{neilfpp} both propose to jointly optimize scene geometry, material, and environment illumination. Such joint optimization in return improves the geometry reconstruction quality. Otherwise, \textbf{NeRO}~\cite{nero} presents a two-stage approach with stage one approximating shading effects and reconstructing object geometry, and stage two recovering the environment lights and the bidirectional reflectance distribution function (BRDF) of the object. In particular, it follows NeuS~\cite{neus} to represent the object surface by an SDF encoded by an MLP but with a different color function using a metalness, a roughness, and an albedo, as in Micro-facet BRDF~\cite{cook1982tog}. Based on NeRO, \textbf{TensoSDF}~\cite{tensosdf} improves the reconstruction performance on objects with arbitrary reflective materials. It proposes a roughness-aware incorporation of the radiance and reflectance fields, and combines the tensorial representation with SDF, enabling more detailed geometry reconstruction and reducing training time.

Finally, unconventional types of input are also explored for reconstruction. \textbf{PAEv3d}~\cite{paev3d} chooses event data and utilizes motion, geometry, and density priors behind event data to impose strong physical constraints to augment NeRF training. A dataset for event data reconstruction with 101 objects is also proposed in PAEv3d.


\subsubsection{Larger-Scene}

Moving on to larger-scene reconstruction, neural RGB-D surface reconstruction by \textbf{NeuralRGB-D}~\cite{azinovic2022cvpr}, as a concurrent work of NeuS~\cite{neus}, also adopts SDF representation for reconstruction. In addition to the RGB supervision, it creates depth supervision when training NeRF. Finally, it is capable of handling noisy pose input with an extra pose refinement module. 
Instead of encoding the entire scene with the MLPs, \textbf{Vox-Surf}~\cite{voxsurf} employs a hybrid architecture that consists of an explicit dense voxel grid with the neural implicit surface representation. The learnable features stored in the voxel grids encode the local geometry, making it possible to recover the entire scene with a smaller MLP. Similar ideas can also be found in \textbf{Instant-NGP}~\cite{instantngp} and \textbf{GO-Surf}~\cite{gosurf}. Specifically, Instant-NGP~\cite{instantngp} introduces a multi-resolution hashing encoding to store voxels in a hash table, enabling much faster ray marching and high-definition resolution with much smaller MLPs. Likewise, GO-Surf~\cite{gosurf} adopts a 4-level feature grid to store the multi-resolution features of the scene geometry. In particular, it introduces grid-based SDF regularization in training and only decodes color in the finest resolution.

To further improve the reconstruction quality, various approaches have been explored. 
\textbf{ManhattanSDF}~\cite{manhattansdf} focuses on boosting reconstruction performance in planar regions. To do so, it introduces additional semantic supervision in training to identify planar regions, and then enforce geometric constraints based on the Manhattan-world assumption. Focusing on textureless regions, \textbf{NeuRIS}~\cite{neuris} introduces adaptive normal supervision with pre-trained single-view normal estimation network and multi-view consistency. Also employing multi-modal supervision signals, \textbf{PMVC}~\cite{pmvc} introduces photometric, depth, normal, semantic, and multi-scale image feature constraints for better multi-view consistency. In addition, correspondence loss based on image features and normal is introduced with Eikonal loss to regularize SDF values in the 3D space. 
Another approach is separating the training of foreground objects from the background. \textbf{H$_2$O-SDF}~\cite{h2osdf}, as an example, proposes a 2-phase learning scheme with phase one focusing on the holistic indoor scene and phase two addressing the specific objects. An object surface field (OSF) is also presented to encapsulate small-scale geometry and high-frequency details on the objects. 
Apart from improving reconstruction, \textbf{Du-NeRF}~\cite{du-nerf} tackles the issue that existing NeRF-based reconstruction methods have unsatisfactory novel view synthesis performance. It employs two separate geometric fields, an SDF field for reconstruction and a density field for novel view synthesis.

In order to alleviate the requirements of depth measurements in training, deep-learning-based depth prediction models are explored. \textbf{MonoSDF}~\cite{monosdf} utilizes the Omnidata~\cite{omnidata} to get depth prediction and train the model with photometric, depth, and normal losses. MonoSDF also investigates 4 different scene representations: dense SDF grid, single MLP, single resolution feature grid with an MLP decoder, and multi-resolution feature grids with an MLP decoder, and discovers that a simple MLP performs the best but is slower to converge while multi-resolution feature grids, in general, can converge fast and capture details, but are less robust to noise and ambiguities in the input images. Furthermore, \textbf{Dong \textit{et al.}}~\cite{dong2023cvpr} also favors the Omnidata~\cite{omnidata} model for depth prediction. However, for speed consideration, it completely discards MLPs and directly stores SDF values in the locally dense yet globally sparse voxel blocks.  In addition to the differentiable volume rendering losses, as in MonoSDF, another dense conditional random field (CRF) is employed to use semantic information for better reconstruction.
Rather than introducing additional depth prediction networks, traditional multi-view stereo methods can also be utilized to acquire the necessary geometry prior. \textbf{NeuralRoom}~\cite{neuralroom}, for example, employs traditional multi-view stereo methods to get an estimated depth map. However, it still relies on pretrained neural networks to compute the normal and its uncertainty.  \textbf{HelixSurf}~\cite{helixsurf}, on the other hand, completely depends on traditional path match multi-view algorithms to predict the per-pixel depth and normal. Then, it presents an iterative intertwined regularization between the neural implicit surface learning and the patch match optimization for the best integration.

Finally, to reconstruct the city-scale environment, \textbf{PC-NeRF}~\cite{pcnerf} proposes a hierarchical spatial partitioning and multi-level scene representation. PC-NeRF designs a parent-child NeRF structure in which parent and child share an MLP. In particular, it divides the entire autonomous vehicle driving environment into large blocks to train the parent NeRF and further divides a block into geometric segments to be represented by child NeRFs.

\subsubsection{Sequential-Input}

When dealing with sequential input, there are commonly two approaches. 
The first one formulates reconstruction with sequential input as a continual learning problem, and it was first explored in continual neural mapping~\cite{cnm2021iccv}. 
Following this idea, \textbf{Hi-Map}~\cite{himap} focuses on indoor reconstruction with a sequence of RGB images. Unlike other monocular reconstruction methods, which use pretrained depth estimation models, it proposes to learn the geometry and appearance features separately and presents multi-resolution factorized feature grids for each decoder.
However, even with the help of continual learning algorithms, the size of a scene that can be represented by a single MLP is limited. \textbf{NISB-Map}~\cite{nisbmap} introduces a scalable mapping framework with multiple MLPs. It uses each MLP to represent a neural implicit surface block, covering a fixed size of a given scene. Such a design significantly reduces memory consumption when reconstructing large-scale environments.
Moving on to outdoor scenes, \textbf{SHINE-Mapping}~\cite{shinemapping} adopts the regularization-based continual learning method to handle the catastrophic forgetting. It favors the SDF surface representation and employs a hybrid scene representation with a learnable octree-based hierarchical feature grid and a globally shared MLP decoder. Also uses a hybrid scene representation, \textbf{RIM}~\cite{rim} stores the implicit features in a multi-resolution voxel map. In particular, it proposes a robot-centric local map whose size is set according to the perceptual range of the sensor to boost training efficiency and curb catastrophic forgetting. The local submaps are later aggregated into a decoupled scalable global map to archive the learned features and maintain constant video memory (VRAM) consumption.

Apart from the continual learning approach, the other one utilizes the recurrent neural network to handle the sequential input. \textbf{NeRFusion}~\cite{nerfusion}, as an example, constructs local feature volume using image features extracted with a convolutional neural network (CNN). Then, the local feature volumes are fused into a global feature volume with a 3D CNN followed by gated recurrent units (GRU). Finally, rendered output are generated with this global feature volume.

%% file: sections/4.2_deformable_recon_v3.tex
Deformable reconstruction mainly accounts for flexible or changing shapes. However, in this subsection, we extend the idea of deformable reconstruction to include 3D reconstructions of dynamic environments as well.
Since deformable NeRF is usually associated with other tasks such as face reconstruction, 3D reconstruction editing, \textit{etc.}, this paper is more concerned with the application of NeRF in autonomous robots. So, the content of this section is mainly investigated in the two directions of \textit{physics-based deformable reconstruction} and \textit{deformation with dynamic sequence}. 
\subsubsection{Physics-based}

Several works have investigated variable reconstruction while incorporating the effects of physical properties, including elastic deformation, kinematic, and dynamic parameters. The inclusion of physical properties makes the reconstructed objects more realistic in terms of deformation and greatly optimizes the interaction with the robot. Physics-based NeRF research can be divided into two categories in terms of representation methods: explicit and implicit. 

For the explicit method, NeRF is usually only applied for static 3D reconstruction of objects or scenes. The NeRF reconstructed models are extracted into meshes, Voxel Grids, or particles, and then physical properties, such as Young's modulus and platonic ratio, are applied to simulate the deformation of the object or the dynamics of the scene. In this approach, the physical simulation is not involved in the design of the loss function in the NeRF training process; it prefers to drive the NeRF reconstruction of the object model as a prelude step to the simulation. \textbf{Neural Impostor}\cite{liu2023neural} uses a hybrid approach to combine explicit mesh and implicit field to achieve more detailed reconstruction and deformation simulation of the physical properties of the soft body based on meshes.\textbf{Pie-NeRF}\cite{feng2023pienerf} adopts a similar idea, except that it extracts discrete points from instant-NGP\cite{instantngp} and uses these discrete points to construct a Voronoi-partition. The simulation of physical properties is based on the Voronoi-partition.

  
In contrast to the explicit approach, the implicit approach takes into account the influence of the physical simulation in the design of the Loss function. They normally require a differentiable simulator and provide the learnable physics parameters as part of the results. \textbf{Chen \textit{et al.}} \cite{chen2022virtual} based on NR-NeRF\cite{nr-nerf:}. It uses discrete points to represent the neural field. It simulates elastic objects with a differentiable particle simulator to learn the parameters of the material. Similarly, the \textbf{PAC-NeRF}\cite{li2023pacnerf} sample points from the voxel NeRF\cite{nr-nerf:}, and use the differentiable material point method (MPM) to achieve the physical simulation. The simulated points will be transferred into the voxel grids again to achieve the later voxel render. In this way, the NeRF is augmented with physics properties. Particularly, PAC-NeRF supports multiple materials such as sand, elastic, and Newtonian fluid.
Several other works focus on the dynamic simulation of rigid objects. \textbf{DANO}\cite{cleach2023differentiable} generates the dynamic-augmented Neural Objects, which contain the object mass and friction properties.




\subsubsection{Deformation with dynamic sequence:}

    Nerf-based reconstruction of deformable objects commonly involves some extent of dynamic scenes, where deformation and reconstruction of objects over time are learned through video. Dynamic reconstruction of the human body or face is also included in this type of research. However, since this paper focuses on NeRf applications in the robotics field, only reconstructions of dynamic scenes or objects are presented in this section. Compared to static scenes, dynamic objects typically have more time-series information, so the key idea of most works is to represent static scenes or objects in the canonical space and then process the time-series information by separate approaches.

    \textbf{D-NeRF}\cite{d-nerf_2021_CVPR} enables end-to-end reconstruction of dynamic scenes and deformable objects from a single RGB video. It inherently learns one network to represent volume density in canonical space, and another deformation network achieves the wrapping from deformation to the canonical space. Similarly, the \textbf{Nerfies}\cite{park2021nerfies} achieves dynamic processing by transforming from deformation to canonical space with MLP named as template NeRF. Additionally, they associate a latent deformation code and an appearance code for each image to assist this transformation. \textbf{HyperNeRF}\cite{park2021hypernerf} extends the canonical space and template NeRF in Nerfies into higher 5D dimensions. In its hyper-canonical space, apart from the spatial pose \((x,y,z)\), there is an MLP that provides other two higher dimensional information \(w_{1},w_{2}\). \textbf{NR-NeRF}\cite{nr-nerf:} also employs a network to represent information such as density and color but differs in that it deals with deformation in the form of rays bending.

    Above approaches prefer to consider the dynamic scene as a whole and transform it directly from deformation to canonical space. Some other works explore separating the dynamic parts from the static parts. The dynamic parts contain time information. And the reconstruction of the whole scene is completed by associating the dynamic and static parts in a reasonable manner.  
    \textbf{Total-Recon}\cite{total-recon} represents the whole dynamic scene as a background Nerf and several dynamic objects Nerf. It changes the pose of the rigid objects in the scene with time-varying, while the non-rigid objects are represented by an additional deformation field. 
    \textbf{STaR}\cite{yuan2021star} holds a similar idea. It completes the reconstruction of the dynamic scene by simultaneously optimizing the parameters of both NeRFs and the relative pose between them.  
    \textbf{Ost \textit{et al.}}\cite{Ost_2021_CVPR} proposed a neural scene graph to represent the scene hierarchically. The nodes in the graph contain a static node and multiple dynamic nodes representing different dynamic objects. Each node contains a local NeRF for a particular object class. The edges of the graph contain the position of the dynamic node relative to the world coordinate system.\(D^{2}\)\textbf{NeRF}\cite{wu2022d2nerf} decouples the static background and dynamic parts of the scene by representing them with two separate NeRFs, and the dynamic parts are also related to the timing. It further adds a shading constraint to avoid overfitting. \textbf{SUDS}\cite{turki2023suds} utilized photometric, geometric, and feature to segment the 2d images into more than two parts: the far-field environment is further segmented, which is represented by a different radiant field. It achieves fast reconstruction of large-scale outdoor dynamic scenes. \textbf{NeRF-DS}\cite{zhiwen2023nerfds} improves the performance of dynamic NeRF reconstruction on reflection conditions. This is achieved by adding masks to dynamic parts and controlling the position and orientation of the dynamic objects surface. 
    
    In addition to the implicit representation of scene and timing sequence variations with networks described above, some work has attempted to represent them explicitly. \textbf{K-Planes}\cite{kplanes_2023_CVPR} and \textbf{HexPlane}\cite{cao2023hexplane} have attempted to represent the canonical space using the feature plane approach with three standard planes. Another three planes were used to learn the timing information. This approach achieves an explicit representation while saving storage space and speeding up computation.

%% file: sections/4.3_recon_eval.tex
\label{sec::recon_eval}
\textbf{Datasets:} When evaluating the above-mentioned methods, a large number of datasets are used. Regarding rigid reconstructions, we focus on the most used three and provide brief descriptions. For \textit{object-level reconstruction}, \textbf{DTU-MVS}~\cite{dtumvs} is the most used dataset. It consists of 80 scenes with each one containing a unique object and 49 to 64 views of the object. All these views have a resolution of 1200$\times$1600 pixels and are taken from a sphere with a radius of 50cm to 65cm. In addition to image views, the DTU-MVS dataset also provides the ground truth geometry and camera intrinsic and extrinsic. 
When it comes to \textit{larger-scene reconstruction}, Replica~\cite{replica} and ScanNet~\cite{scannet} are more favored. In particular, \textbf{Replica}~\cite{replica} is a dataset of high-fidelity replicates of 18 room-scale indoor spaces. For each space, it provides ground truth dense geometry, high-resolution HDR textures, reflectors, and semantic class and instance annotations. Moreover, 6 spaces are the same space with its contents differently arranged, while the other 12 spaces are completely semantically different indoor scenes. 
\textbf{ScanNet}~\cite{scannet}, on the other hand, provides 2.5 million views across 1,513 RGB-D video scans of 707 indoor rooms. It provides pseudo ground truth 3D camera poses and surface reconstructions generated by BundleFusion~\cite{bundlefusion}, along with annotated instance-level semantic segmentations. 

Apart from these three, a wide range of datasets have been explored to evaluate reconstruction performance. For example, datasets that are popular in novel view synthesis tasks like \textit{Synthetic Objects}\cite{nerf-c, nerf-j}, \textit{LLFF}~\cite{llff}, \textit{Tanks \& Temples}~\cite{tankstemples} and \textit{BlendedMVS}~\cite{blendedmvs}; datasets with event inputs like \textit{EventNeRF}~\cite{eventnerf} and \textit{PAEv3d}~\cite{paev3d}; and datasets that include reflective objects, such as \textit{NeILF-Synthetic}~\cite{neilf}, \textit{NeILF-HDR}~\cite{neilfpp}, \textit{Glossy-Blender/Real}~\cite{nero}, and \textit{TensoSDF-Synthetic}~\cite{tensosdf}. Furthermore, to evaluate reconstruction performance on more general and practical scenarios, larger-scale datasets are also selected in different papers. For example, \textit{10 Synthetic Scenes}~\cite{azinovic2022cvpr}, \textit{7-Scenes}~\cite{7scenes}, \textit{SceneNet}~\cite{scenenet} for indoor scenes; \textit{EPFL}~\cite{epfl}, \textit{Mai City}~\cite{maicity}, \textit{KITTI}~\cite{kitti} and \textit{KITTI-360}~\cite{kitti360} for outdoor scenes ; and \textit{ETH3D}~\cite{eth3d} and \textit{RealEstate10K}~\cite{realestate10k} for both indoor and outdoor scenes. 

For deformable reconstruction, most of the works focus on the synthesis of new images of dynamic scenes, using the D-NeRF dataset\cite{d-nerf_2021_CVPR} or Plenoptic Dataset\cite{li2022neuralPlenopticDataset} for evaluation. Deformable reconstruction or dynamics reconstruction in these works prefers to present it as a function; therefore, no unified dataset and metrics are currently available for the deformable reconstruction.  

\textbf{Evaluation Metrics:} We provide an overview of popular metrics, namely Accuracy (Acc.), Completion (Comp.), Chamfer-$L1$ distance (C-$l_1$), Precision (Prec.), Recall (also known as Completion Ratio, C.R.), F-Score, and Normal Consistency (N.C.), when evaluating the reconstruction performance. Assume $P_{pred}$ and $P_{gt}$ are predicted and ground truth point clouds, we have $\text{C-}l_1 := (\text{Acc.} + \text{Comp.})/2$, $\text{F-score} := (2\cdot \text{Prec.}\cdot \text{Recall}) / (\text{Prec.}+ \text{Recall})$, and:
\begin{equation}
    \text{Acc.} := \frac{1}{|P_{pred}|}\sum_{\bm{p}\in P_{pred}} \text{min}_{\bm{q}\in P_{gt}}||\bm{p}-\bm{q}||
\end{equation}
\vspace{-2ex}
\begin{equation}
    \text{Comp.} := \frac{1}{|P_{gt}|}\sum_{\bm{q}\in P_{gt}} \text{min}_{\bm{p}\in P_{pred}}||\bm{p}-\bm{q}||
\end{equation}
\vspace{-2ex}
\begin{equation}
    \text{Prec.} := \frac{1}{|P_{pred}|}\sum_{\bm{p}\in P_{pred}} ((\text{min}_{\bm{q}\in P_{gt}}||\bm{p}-\bm{q}||)<\tau)
\end{equation}
\vspace{-2ex}
\begin{equation}
    \text{Recall} := \frac{1}{|P_{gt}|}\sum_{\bm{q}\in P_{gt}} ((\text{min}_{\bm{p}\in P_{pred}}||\bm{p}-\bm{q}||)<\tau)
\end{equation}
\vspace{-2ex}
\begin{equation}
    \text{N.C.} := \frac{1}{2}(\frac{1}{|P_{pred}|}\sum_{\bm{p}\in P_{pred}} \bm{n_{\bm{p}}}^T\bm{n_{\bm{q}}} + \frac{1}{|P_{gt}|}\sum_{\bm{q}\in P_{gt}} \bm{n_{\bm{p}}}^T\bm{n_{\bm{q}}})
\end{equation}

\begin{table*}[t]
\caption{Reconstruction quality in terms of C-$l_1$ [mm]$\downarrow$ of Rigid Objects on DTU-MVS dataset~\cite{dtumvs}, results are taken from their original papers. \textbf{Bold} means the best performance and \underline{underline} means the second best.}
\label{tab:recon-obj}
\scriptsize
\begin{tabular*}{\textwidth}{@{\extracolsep\fill}p{2.2cm}|ccccccccccccccc|c}
  \toprule
  Scans & 24 & 37 & 40 & 55 & 63 & 65 & 69 & 83 & 97 & 105 & 106 & 110 & 114 & 118 & 122 & Avg. \\
  \midrule
  COLMAP$^a$\cite{colmap-sfm} & 0.81 & 2.05 & 0.73 & 1.22 & 1.79 & 1.58 & 1.02 & 3.05 & 1.40 & 2.05 & 1.00 & 1.32 & 0.49 & 0.78 & 1.17 & 1.36 \\ 
  IDR$^b$\cite{idr} & 1.63 & 1.87 & 0.63 & 0.48 & 1.04 & 0.79 & 0.77 & 1.33 & 1.16 & 0.76 & 0.67 & 0.90 & 0.42 & 0.51 & 0.53 & 0.90 \\ 
  \midrule
  UNISURF\cite{unisurf} & 
  1.32 & 1.36 & 1.72 & 0.44 & 1.35 & 0.79 & 0.80 & 1.49 & 1.37 & 0.89 & 0.59 & 1.47 & 0.46 & 0.59 & 0.62 & 1.02 \\ 
  Discrete-UNISURF\cite{jiang2023iccv} & 0.85 & 0.95 & 1.00 & 0.38 & 1.25 & 0.59 & 0.69 & 1.36 & 1.19 & 0.71 & 0.52 & 1.15 & 0.42 & 0.48 & 0.50 & 0.80 \\ 
  NeuS\cite{neus} & 1.00 & 1.37 & 0.93 & 0.43 & 1.10 & 0.65 & 0.57 & 1.48 & 1.09 & 0.83 & 0.52 & 1.20 & 0.35 & 0.49 & 0.54 & 0.84 \\ 
  NeuS$^b$\cite{neus} & 0.83 & 0.98 & 0.56 & 0.37 & 1.13 & 0.59 & 0.60 & 1.45 & 0.95 & 0.78 & 0.52 & 1.43 & 0.36 & 0.45 & 0.45 & 0.77 \\ 
  Discrete-NeuS\cite{jiang2023iccv} & 0.71 & 0.90 & 0.68 & 0.38 & 1.00 & 0.60 & 0.58 & 1.40 & 1.17 & 0.78 & 0.52 & 1.07 & 0.32 & 0.43 & 0.45 & 0.73 \\ 
  VolSDF\cite{volsdf} & 1.14 & 1.26 & 0.81 & 0.49 & 1.25 & 0.70 & 0.72 & 1.29 & 1.18 & 0.70 & 0.66 & 1.08 & 0.42 & 0.61 & 0.55 & 0.86 \\ 
  Geo-NeuS\cite{geoneus} & \underline{0.38} & \textbf{0.54} & \textbf{0.34} & \underline{0.36} & 0.80 & \textbf{0.45} & \textbf{0.41} & \underline{1.03} & \textbf{0.84} & \textbf{0.55} & \textbf{0.46} & \textbf{0.47} & \textbf{0.29} & \textbf{0.36} & \textbf{0.35} & \textbf{0.51} \\ 
  HF-NeuS\cite{hfneus} & 0.76 & 1.32 & 0.70 & 0.39 & 1.06 & 0.63 & 0.63 & 1.15 & 1.12 & 0.80 & 0.52 & 1.22 & 0.33 & 0.49 & 0.50 & 0.77 \\ 
  NeuralWarp\cite{neuralwarp} & 0.49 & 0.71 & 0.38 & 0.38 & 0.79 & 0.81 & 0.82 & 1.20 & 1.06 & 0.68 & 0.66 & 0.74 & 0.41 & 0.63 & 0.51 & 0.68 \\ 
  Discrete-NeuralWarp\cite{jiang2023iccv} & 0.49 & 0.68 & 0.37 & \underline{0.36} & \underline{0.73} & 0.76 & 0.77 & 1.17 & 1.10 & 0.67 & 0.62 & 0.65 & 0.36 & 0.57 & 0.49 & 0.65 \\ 
  LoD-NeuS\cite{lodneus} & 0.69 & 0.88 & 0.47 & 0.42 & 0.85 & 0.94 & 0.59 & \textbf{0.80} & 1.31 & 0.64 & 0.61 & 1.27 & \textbf{0.29} & 0.64 & 0.38 & 0.72 \\
  LoD-NeuS$^b$\cite{lodneus} & 0.65 & 0.91 & 0.37 & 0.48 & 1.05 & 0.87 & 0.82 & 1.22 & 0.95 & 0.69 & 0.56 & 1.30 & 0.42 & 0.58 & 0.57 & 0.76 \\ 
  GenS\cite{gens} & 0.66 & 1.01 & 0.71 & 0.43 & 1.06 & 0.99 & 0.73 & 1.43 & 1.18 & 0.78 & 0.64 & 0.93 & 0.38 & 0.54 & 0.54 & 0.80 \\
  GenS$^c$\cite{gens} & 0.55 & 0.71 & 0.39 & 0.38 & 0.79 & 0.65 & 0.57 & 1.29 & 0.96 & 0.64 & 0.49 & 0.59 & 0.33 & 0.44 & 0.45 & 0.62 \\
  Voxurf\cite{voxurf} & 0.65 & 0.74 & 0.39 & \textbf{0.35} & 0.96 & 0.64 & 0.85 & 1.58 & 1.01 & 0.68 & 0.60 & 1.11 & 0.37 & 0.45 & 0.47 & 0.72 \\
  PSDF\cite{psdf} & \textbf{0.36} & \underline{0.60} & \underline{0.35} & \underline{0.36} & \textbf{0.70} & 0.61 & \underline{0.49} & 1.11 & \underline{0.89} & \underline{0.60} & \underline{0.47} & \underline{0.57} & \underline{0.30} & \underline{0.40} & \underline{0.37} & \underline{0.55} \\
  \midrule
  Vox-Surf\cite{voxsurf} & 0.72 & 1.15 & 0.51 & \textbf{0.35} & 1.09 & \underline{0.58} & 0.59 & 1.35 & 0.91 & 0.77 & \textbf{0.46} & 1.09 & 0.35 & 0.42 & 0.43 & 0.72 \\ 
  MonoSDF-MLP\cite{monosdf} & 0.83 & 1.61 & 0.65 & 0.47 & 0.92 & 0.87 & 0.87 & 1.30 & 1.25 & 0.68 & 0.65 & 0.96 & 0.41 & 0.62 & 0.58 & 0.84 \\ 
  MonoSDF-Grids\cite{monosdf} & 0.66 & 0.88 & 0.43 & 0.40 & 0.87 & 0.78 & 0.81 & 1.23 & 1.18 & 0.66 & 0.66 & 0.96 & 0.41 & 0.57 & 0.51 & 0.73 \\ 
  \midrule
  \midrule
  SparseNeuS$^d$\cite{sparseneus} & 1.68 & 3.06 & 2.25 & 1.10 & 2.37 & 2.18 & 1.28 & 1.47 & 1.80 & 1.23 & 1.19 & 1.17 & 0.75 & 1.56 & 1.55 & 1.64 \\ 
  SparseNeuS$^{cd}$\cite{sparseneus} & 1.29 & \textbf{2.27} & 1.57 & \underline{0.88} & 1.61 & \underline{1.86} & 1.06 & \underline{1.27} & 1.42 & 1.07 & \underline{0.99} & \underline{0.87} & \underline{0.54} & 1.15 & 1.18 & \underline{1.27} \\ 
  VolRecon$^d$\cite{volrecon} & \underline{1.20} & 2.59 & \underline{1.56} & 1.08 & \underline{1.43} & 1.92 & 1.11 & 1.48 & 1.42 & 1.05 & 1.19 & 1.38 & 0.74 & 1.23 & 1.27 & 1.38 \\ 
  GenS$^d$\cite{gens} & 1.45 & 2.77 & 1.69 & 0.97 & 1.54 & 1.90 & \underline{1.03} & 1.49 & \underline{1.36} & \underline{0.97} & 1.07 & 0.97 & 0.62 & \underline{1.14} & \underline{1.16} & 1.34 \\
  GenS$^{cd}$\cite{gens} & \textbf{0.91} & \underline{2.33} & \textbf{1.46} & \textbf{0.75} & \textbf{1.02} & \textbf{1.58} & \textbf{0.74} & \textbf{1.16} & \textbf{1.05} & \textbf{0.77} & \textbf{0.88} & \textbf{0.56} & \textbf{0.49} & \textbf{0.78} & \textbf{0.93} & \textbf{1.03} \\
  
  \bottomrule
\end{tabular*}
\vspace{-2ex}
\begin{flushleft}
\scriptsize{
$^a$COLMAP results are achieved by trim=0, taken from IDR~\cite{idr} \\ 
$^b$Methods trained with instance masks; $^c$Results obtained with per-scene finetuning. \\
$^d$Trained with only sparse views: 3 views for SparseNeus~\cite{sparseneus} and VolRecon~\cite{volrecon}, 4 views for GenS~\cite{gens}. \\
}
\end{flushleft}
\vspace{-6ex}
\end{table*}

\begin{table*}
\caption{Reconstruction quality of room-scale environments on Replica~\cite{replica} and ScanNet~\cite{scannet} datasets. Note the unit for Acc. Comp. and C-$l_1$ is [cm], the threshold $\tau$ for Prec., Recall, and F-score is 5cm. \textbf{$-$} means the original paper has no report for the benchmark.}
\label{tab:recon-room}
\scriptsize
\begin{tabular*}{\textwidth}{@{\extracolsep\fill}p{2cm}|cccccc|ccccc}
  \toprule
  \multirow{2}{*}{Methods} & 
  \multicolumn{6}{c|}{ScanNet} 
  & \multicolumn{5}{c}{Replica} \\
  \cline{2-7} \cline{8-12}
  & Acc.$\downarrow$ & Comp.$\downarrow$ & Prec.$\uparrow$ & Recall$\uparrow$ & F-score$\uparrow$ & C-$l_1$$\downarrow$
  & Acc.$\downarrow$ & Comp.$\downarrow$ & F-score$\uparrow$ & C-$l_1$$\downarrow$ & N.C.$\uparrow$\\
  \midrule
  COLMAP\cite{colmap-sfm} & 0.047$^{abcd}$ & 0.235$^{abcd}$ & 0.711$^{abcd}$ & 0.441$^{abcd}$ & 0.537$^{abcd}$ & 0.141$^c$ 
  & - & - & - & - & - \\  
  NeuS\cite{neus} & 0.179$^{abcd}$ & 0.208$^{abcd}$ & 0.313$^{abcd}$ & 0.275$^{abcd}$ & 0.291$^{abcd}$ & 0.194$^c$ 
  & 0.162 & 0.296 & 0.194$^e$ & 0.229$^e$ & 0.754$^e$ \\  
  VolSDF\cite{volsdf} & 0.414$^{abcd}$ & 0.120$^{abcd}$ & 0.321$^{abcd}$ & 0.394$^{abcd}$ & 0.346$^{abcd}$ & 0.267$^c$ 
  & 0.135$^e$ & 0.301$^e$ & 0.339$^e$ & 0.218$^e$ & 0.747$^e$ \\ 
  \midrule
  Jiang \textit{et al.}~\cite{jiang2023iccv} & - & - & 0.794 & 0.750 & 0.770 & 0.039 & - & - & 0.902 & 0.028 & 0.939 \\ 
  Neural RGB-D\cite{azinovic2022cvpr} & - & - & - & - & - & - 
  & \textbf{0.010}$^e$ & 0.245$^e$ & 0.847$^e$ & 0.127$^e$ & 0.934$^e$ \\ 
  Go-Surf\cite{gosurf} & - & - & - & - & - & - 
  & 0.012$^e$ & \underline{0.012}$^e$ & \underline{0.990}$^e$ & \underline{0.012}$^e$ & \underline{0.972}$^e$ \\
  Instant-NGP\cite{instantngp} & - & - & - & - & - & -  
  & 0.333$^e$ & 1.115$^e$ & 0.158$^e$ & 0.729$^e$ & 0.547$^e$ \\
  Manhattan SDF\cite{manhattansdf} & 0.072 & 0.068 & 0.621 & 0.586 & 0.602 & 0.070
  & - & - & - & - & -  \\ 
  NeuRIS\cite{neuris} & 0.05$^{bcd}$ & 0.05$^{bcd}$ & 0.71$^{bcd}$ & 0.67$^{bc}$ & 0.69$^{bcd}$ & 0.050$^c$ 
  & - & - & - & - & -  \\ 
  PMVC\cite{pmvc} & 0.038 & \underline{0.039} & \underline{0.815} & \textbf{0.774} & \underline{0.794} & \underline{0.038} 
  & - & - & 0.900 & 0.027 & 0.941  \\ 
  MonoSDF-MLP\cite{monosdf} & \underline{0.035}$^{bcd}$ & 0.048$^{bcd}$ & 0.799$^{bcd}$ & 0.681$^{bcd}$ & 0.733 & 0.042
  & - & - & 0.862 & 0.029 & 0.921 \\ 
  MonoSDF-Grids\cite{monosdf} & - & - & - & - & 0.626 & 0.064
  & - & - & 0.859 & 0.032 & 0.909 \\ 
  Dong \textit{et al.}\cite{dong2023cvpr} & 0.042 & 0.056 & 0.751 & 0.678 & 0.710 & 0.049 
  & - & - & - & - & -  \\ 
  HelixSurf\cite{helixsurf} & 0.038 & 0.044 & 0.786 & 0.727 & 0.755 & 0.041 
  & - & - & - & - & -  \\ 
  H$_2$O-SDF\cite{h2osdf} & \textbf{0.032}$^f$ & \textbf{0.038}$^f$ & \textbf{0.836}$^f$ & \underline{0.770}$^f$ & \textbf{0.801}$^f$ & \textbf{0.035}$^f$ 
  & - & - & - & - & -  \\ 
  Hi-Map\cite{himap} & - & - & - & - & - & -
  & 0.060 & 0.074  & - & - & - \\ 
  Du-NeRF\cite{du-nerf} & - & - & - & - & - & - 
  & \underline{0.011} & \textbf{0.011} & \textbf{0.991} & \textbf{0.011} & \textbf{0.975} \\ 
  \bottomrule
\end{tabular*}
\vspace{-2ex}
\begin{flushleft}
\scriptsize{
$^a$Results from ManhattanSDF~\cite{manhattansdf} paper; 
$^b$Results from HelixSurf~\cite{helixsurf} paper; \\
$^c$Results from PMVC~\cite{pmvc} paper;  
$^d$Results from H$_2$O-SDF~\cite{h2osdf} paper; 
$^e$Results from Du-NeRF~\cite{du-nerf} paper; \\
$^f$H$_2$O-SDF~\cite{h2osdf} results are re-computed for the same 4 scenes used by other methods, using the per-scene results from their paper. 
}
\end{flushleft}
\vspace{-6ex}
\end{table*}

\textbf{Results Summary:} Finally, we summarize the reported results from the above-mentioned papers.
For the DTU-MVS~\cite{dtumvs} dataset, 15 challenging scans, out of 80, selected by IDR~\cite{idr} are used for evaluation. We show the summarized results in Table~\ref{tab:recon-obj}. Then, For Replica~\cite{replica}, all the evaluations are carried out in office 0-4 and room 0-2. Finally, for ScanNet~\cite{scannet}, there is a huge difference in scene selections from paper to paper. To best compare the performance, we choose the most followed scene selections used in ManhattanSDF~\cite{manhattansdf} (scene 0050, 0084, 0580, 0616). The results from Replica and ScanNet are shown in Table~\ref{tab:recon-room}.
Note that there are inevitable result conflicts between different papers. To ensure a fair comparison, we favor the values reported in the original paper. When the paper does not provide a conflicting result, we favor the most consistent one across different literature. Such results are marked with specific superscript regarding its source. Finally, when a method has no consistent results at all,  we report the values with a lower numerical precision that is mostly agreed upon.

%% file: sections/7_segmentation.tex
NeRF has also been widely used for different segmentation problems. In the context of autonomous robots, segmentation is crucial for tasks such as object recognition, scene understanding, and navigation. For example, a robot might need to segment an image captured by its camera into different objects to identify and interact with them, or it might need to segment a map of its environment into navigable and non-navigable areas. Effective segmentation allows robots to process and interpret complex data, enabling them to perform tasks more efficiently and accurately in dynamic and unstructured environments. We classify works of segmentation with NeRF into three main categories: 
\textit{segmentation interpolation}, \textit{multi-view consistent 2D segmentation rendering}, and \textit{domain adaptation}.

\subsection{Segmentation Interpolation}
One of the most important features of NeRF is the capability of interpolation --- learning dense relations from sparse training signals. This not only increases the training efficiency but also alleviates annotation burdens. NeRF models can propagate a set of very sparse segmentation labels to the whole image space by training a semantic field. The first work that combines segmentation formulation with NeRF is \textbf{Semantic-NeRF} \cite{zhi2021place}, where the authors extend the original NeRF \cite{nerf-j} to jointly learn appearance, geometry, and semantics. The model structure of Semantic-NeRF is constructed by appending a segmentation renderer head before injecting viewing directions into the MLP. The training objective consists of a photometric loss that is based on RGB values and a semantic loss calculated by cross-entropy. With the modified model structure and the new training loss, Semantic-NeRF is able to learn dense 2D segmentations for novel-view images from sparse labeled pixels. The authors show that the model can perform several semantic label denoising tasks, including learning from labels with pixel-wise noise, learning from labels with region-wise noise, semantic super-resolution, label propagation, and multi-view semantic fusion. \textbf{iLabel} \cite{zhi2021ilabel} uses the Semantic-NeRF in a labeling system, where the model allows the users to focus on interactions to achieve efficient labeling. It has been proved that a room or similar scene can be correctly labeled into 10+ semantic classes with only a few tens of clicks. Similar to Semantic-NeRF, \textbf{SegNeRF} \cite{zarzar2022segnerf} integrates a semantic field along with the usual radiance field to perform 3D part segmentation from a few images. Being different from Semantic-NeRF, \textbf{NeSF} \cite{nesf} proposes to learn the semantic rendering from a 3D semantic feature grid, which is translated from the density grid through a 3D UNet. In NeSF, the density grid is extracted from a pretrained NeRF model. The learning process from the semantic feature grid to the semantic rendering is performed by a semantic MLP. In contrast to the first-order training objective in Semantic-NeRF, \textbf{JacobiNeRF} \cite{xu2023jacobinerf} proposes a model that encodes semantic correlations between scene points, regions, and entities. The correlation encoding is achieved by using an explicit regularization to the learning dynamics by aligning the Jacobians of highly correlated entities. The proposed model can be useful for label propagation and is superior to the Semantic-NeRF. One limitation of Semantic-NeRF and similar works is they need to train one specific NeRF model for each scene. To improve the generalization of NeRF models when learning semantics, \textbf{Semantic Ray} \cite{liu2023semantic} trains one unified model on multiple scenes and generalizes to unseen scenes. In Semantic Ray, a novel cross-reprojection attention (CRA) module is proposed to construct a refined 3D contextual space for query rays based on convolutional features of the sampled pixels in source views. The CRA module provides rich contextual features for semantic segmentation while avoiding performing dense attention over multi-view reprojected rays. To generate the semantic logits for each query point, a semantic-aware weight network is proposed to rescore the significance of each source view. \textbf{GNeSF} \cite{chen2024gnesf} improves the generalization by taking in multi-view image features and semantic maps as the inputs instead of only spatial information to avoid overfitting to scene-specific geometric and semantic information. A novel soft voting mechanism is proposed to aggregate the 2D semantic information from different views for each 3D point.

\subsection{Multi-view Consistency}
The learning process of 3D fields in NeRF brings multi-view consistency across different views, which benefits several 2D rendering tasks, e.g., instance segmentation and panoptic segmentation. \textbf{SUDS} \cite{turki2023suds} builds a dynamic large-scale urban NeRF by factorizing the scene into three separate hash tables to encode \textit{static}, \textit{dynamic}, and \textit{far-field} radiance fields. To facilitate the construction of the desired NeRF, SUDS makes use of several input data, including RGB images, sparse LiDAR, self-supervised 2D descriptors (from DINO \cite{caron2021emerging}), and 2D optical flow. \textbf{Instance-NeRF} \cite{liu2023instance} learns consistent 3D instance segmentation of a given scene using 2D labels predicted by Mask2Former \cite{cheng2022masked}. However, the predictions from Mask2Former are independent from frame to frame and may lose the correspondence. To achieve consistency across different frames, the authors propose a NeRF-RCNN (extended from NeRF-RPN \cite{hu2023nerf}) that takes the extracted radiance and density field of the pre-trained NeRF as inputs and outputs 3D masks for each detected object. The predicted coarse 3D mask is then projected into image space and the projected mask is used to correct the instance in the mask predicted by Mask2Former. The 2D labels become consistent across frames after correction and are used to train the Instance-NeRF model. Similar to Instance-NeRF, \textbf{Panoptic Lifting} \cite{siddiqui2023panoptic} aims at rendering consistent 2D panoptic images by leveraging NeRF. The model only requires machine-generated 2D masks, which might be inconsistent across frames. Panoptic Lifting adopts several strategies to improve the performance, including (1) applying a segment consistency loss; (2) performing a test-time augmentation; (3) using the bounded segmentation logits; and (4) blocking the gradients from the semantic branch to the geometry branch. Nevertheless, there are two major drawbacks \cite{bhalgat2023contrastive} in Panoptic Lifting: (1) In every gradient computation, the model needs to solve a linear assignment problem using Hungarian Matching. The cost of this increases cubically with the number of identifiers. (2) The label space should be enough to accommodate a potentially large number of objects. To overcome those limitations in Panoptic Lifting, \textbf{Contrastive Lift} \cite{bhalgat2023contrastive} achieves multi-view consistency across frames by using a slow-fast clustering objective, which also eliminates the requirement for an upper bound on the number of objects existing in the scene. \textbf{PNF} \cite{kundu2022panoptic} uses an MLP to model each object and the MLP is instance-specific and thus can be smaller and faster than previous object-aware approaches, while still leveraging category-specific priors incorporated via meta-learned initialization. \textbf{Nerflets} \cite{zhang2023nerflets} uses a group of \textit{nerflets} to represent each object instance. A nerflet is defined as a small local MLP that focuses only on a small portion of the scene determined by its influence function, which is an analytic radial basis function based on scaled anisotropic multivariate Gaussians. Results from individual nerflets are mixed based on the influence values. \textbf{Panoptic NeRF} \cite{fu2022panoptic} aims for
obtaining per-pixel 2D semantic and instance labels from
easy-to-obtain coarse 3D bounding primitives, which cover the full scene in the form of cuboids, ellipsoids, and extruded polygons. The authors of Panoptic NeRF propose a semantically-guided geometry optimization to improve the underlying geometry, and a joint geometry and semantic optimization to further resolve label ambiguity at intersection regions of the 3D bounding primitives. \textbf{PlanarNeRF}~\cite{chen2023planarnerf} proposes a 3D plane segmentation model based on NeRF. The work formulates the 3D plane detection as a 2D plane rendering problem and introduces a plane rendering branch to the NeRF model. An efficient RANSAC algorithm is used to estimate local planes. A global memory bank is proposed to perform the tracking to maintain consistency among planes estimated for each frame.

\subsection{Domain Adaptation}
The potential of using NeRF in domain adaptation is first explored by \cite{liu2023unsupervised}, where a NeRF-based model and a semantic segmentation model are jointly trained to achieve continual adaptation in a series of scenes. The Semantic-NeRF \cite{zhi2021place} is used as the radiance field model, and the DeepLabv3 \cite{chen2017rethinking} with ResNet-101 \cite{he2016deep} is used as the segmentation model. A pretrained segmentation model $f_{\theta_0}$ is assumed to be available. The goal of \cite{liu2023unsupervised} is to adapt $f_{\theta_0}$ across a sequence of NN scenes. For each scene $S_i$, the objective is to find a set of weights $\theta_i$ of the segmentation model, starting from $\theta_{i-1}$, such that the performance of $f_{\theta_i}$ on $S_i$ is higher than that of $f_{\theta_{i-1}}$. More specifically, the authors train a Semantic-NeRF model for each scene $S_i$, where the semantic label is provided by $f_{\theta_{i-1}}$. Then the rendered segmentation images (pseudo labels) from the trained Semantic-NeRF are further used to train $f_{\theta_i}$. The pseudo labels from the Semantic-NeRF show multi-view consistency, which constitutes an important prior that can be exploited to guide the adaptation of the segmentation model.

\subsection{Datasets and Evaluations}
\textbf{Datasets:} Different datasets are used to evaluate the segmentation performance of NeRF-based models. All the datasets can be categorized into three classes: \textit{object-level}, \textit{room-level}, and \textit{outdoor-level}. The most commonly used room-level datasets are \textbf{Replica}~\cite{replica} and \textbf{ScanNet}, which have already been introduced in Section~\ref{sec::recon_eval}. Other than these, there exist several datasets for synthetic room-level scenes, such as \textbf{HyperSim}~\cite{roberts2021hypersim} and \textbf{3D-FRONT}~\cite{fu20213d}.

For object-level datasets, \textbf{ToyBox5} and \textbf{ToyBox13} are two challenging datasets proposed in NeSF~\cite{nesf}, which mimic scenes of children's bedrooms using objects from ShapeNet~\cite{chang2015shapenet}. There are 4-12 randomly-selected objects randomly-placed in each scene. Additionally, 300 independently sampled camera poses are used to render input frames.
\textbf{PartNet}~\cite{mo2019partnet} dataset and \textbf{ShapeNet}~\cite{chang2015shapenet} are involved in evaluating the object-level segmentation performance of NeRF-based models. PartNet dataset contains a collection of object meshes that are divided into parts, along with their corresponding part segmentation labels. However, the PartNet dataset lacks the RGB information for the mesh, being not suitable for NeRF-based settings. To overcome this issue, SegNeRF~\cite{zarzar2022segnerf} finds correspondences between the PartNet dataset and the ShapeNet dataset and extracts the color information for those matched object meshes from the ShapeNet dataset. 



For outdoor scenes, \textbf{KITTI}~\cite{geiger2012we} and \textbf{KITTI-360}~\cite{liao2022kitti} are representative datasets for evaluating segmentation performance in large-scale outdoor environments. The KITTI semantic segmentation benchmark consists of 200 training and 200 test images from the KITTI Stereo and Flow Benchmark 2015. These images are thoroughly annotated for semantic segmentation, with metrics conforming to the Cityscapes~\cite{cordts2016cityscapes} Dataset standards. KITTI-360 is an extensive dataset that enhances the original KITTI dataset by offering a more complete view for semantic and instance segmentation tasks. It provides high-fidelity 360-degree sensory data, including fisheye images and push broom laser scans. In addition, it includes precise geo-localized vehicles and camera poses. The dataset is designed to support a variety of tasks such as 2D and 3D semantic segmentation, instance segmentation, and panoptic segmentation.

\textbf{Evaluation Metrics:} Several metrics are used to evaluate semantic segmentation performance, including the mean of \textbf{I}ntersection over Union (mIoU), \textbf{A}ccuracy, \textbf{P}recision, and \textbf{R}ecall. The definitions for those additional metrics are $\mathbf{I} = \frac{|TP|}{|TP| + |FP| + |FN|}, \mathbf{A} = \frac{|TP|+|TN|}{|TP|+|TN| + |FP| + |FN|}, \mathbf{P} = \frac{|TP|}{|TP| + |FP|}$, and $\mathbf{R} = \frac{|TP|}{|TP| + |FN|}$, where $TP, TN, FP$ and $FN$ are true positive, true negative, false positive, and false negative, respectively.

\begin{table}[t]
\caption{Segmentation results for outdoor datasets.}
\footnotesize
\begin{tabular}{p{2cm}|ccp{0.7cm}|cc }
\toprule
& \multicolumn{3}{c}{KITTI-360} & \multicolumn{2}{c}{KITTI} \\
  Methods  & mIoU$\uparrow$ & Acc.$\uparrow$ & PQ$\uparrow$ & mIoU$\uparrow$ & PQ$\uparrow$ \\
  \midrule
NeRF~\cite{nerf-j} + PSPNet~\cite{zhao2017pyramid} &  53.01 &  - &  - &  - & -  \\
FVS~\cite{riegler2020free} + PSPNet~\cite{zhao2017pyramid} & 67.08 & - & - & - & - \\
PBNR~\cite{kopanas2021point} + PSPNet~\cite{zhao2017pyramid} & 65.07 & - & - & - & - \\
Mip-NeRF~\cite{mipnerfrgbd} + PSPNet~\cite{zhao2017pyramid} & 51.15 & - & - & - & - \\
DeepLab~\cite{chen2017deeplab} on GT RGB & - & - & - & \underline{49.9} & \underline{43.2} \\
DeepLab~\cite{chen2017deeplab} on NeRF rendered RGB & - & - & - & 32.1 & 24.9 \\
PNF~\cite{kundu2022panoptic} & 74.28 & - & - & \textbf{56.5} & \textbf{45.9}\\
Nerflets~\cite{zhang2023nerflets} & 75.07 & - & - & - & - \\
3D-2D CRF~\cite{liao2022kitti} & \underline{79.5} & 92.8 & 62.2$^a$/ 60.7$^b$/ 63.0$^c$ & - & - \\
Panoptic NeRF~\cite{fu2022panoptic} & \textbf{81.1} & 93.2 & 64.4$^a$/ 61.9$^b$/ 65.8$^c$ & - & -\\ 
\bottomrule
\end{tabular}
\vspace{-2ex}
\begin{flushleft}
\footnotesize{$^a$For all classes; $^b$For things class only; $^c$For stuff class only.}
\end{flushleft}
\label{tab:outdoor-seg}
\vspace{-6ex}
\end{table}

\begin{table*}
\caption{Segmentation results for room-level datasets.}
\footnotesize
\begin{tabular}{c|ccc|ccc|ccc}
\toprule
& \multicolumn{3}{c}{Replica} & \multicolumn{3}{c}{ScanNet} & \multicolumn{3}{c}{HyperSim}\\
  Methods  & mIoU$\uparrow$ & Acc.$\uparrow$ & $\text{PQ}^{\text{scene}}\uparrow$ & mIoU$\uparrow$ & Acc.$\uparrow$ & $\text{PQ}^{\text{scene}}\uparrow$ & mIoU$\uparrow$ & Acc.$\uparrow$ & $\text{PQ}^{\text{scene}}\uparrow$\\
  \midrule
Mask2Former~\cite{cheng2022masked} & 52.4 & - & - & 46.7 & - & - & 53.9 & - & - \\
Semantic-NeRF~\cite{zhi2021place} & 58.5 &  - &  - &  59.2 & - & - & \underline{58.9} & - & - \\
DM-NeRF~\cite{wang2022dm} & 56.0 & - & 44.1 & 49.5 & 41.7 & - & 57.6 & 51.6 & - \\
PNF~\cite{kundu2022panoptic} & 51.5 & - & 41.1 & 53.9 & - & 48.3 & 50.3 & - & 44.8 \\
PNF+GT Bounding Box & 54.8 & - & 52.5 & 58.7 & - & 54.3 & 58.7 & - & 47.6 \\
Panoptic Lifting~\cite{siddiqui2023panoptic} & 67.2 & - & \underline{57.9} & 65.2 & - & 58.9 & \textbf{67.8} & - & 60.1 \\
Contrastive Lift (Vanilla)~\cite{bhalgat2023contrastive} & - & - & 57.8 & - & - & \underline{60.5} & - & - & \underline{60.9}\\
Contrastive Lift (SF$^a$)~\cite{bhalgat2023contrastive} & - & - & \textbf{59.1} & - & - & \textbf{62.3} & - & - & \textbf{62.3} \\
MVSNeRF$^b$~\cite{chen2021mvsnerf} + Semantic head & 23.4 & 33.7 & - & 39.8 & 46.0 & - & - & - & - \\
NeuRay$^b$~\cite{liu2022neural} + Semantic head & 35.9 & 44.0 & - & 51.0 & 57.1 & - & - & - & - \\
Semantic Ray$^b$~\cite{liu2023semantic} & 41.6 & 47.2 & - & 57.2 & 62.6 & - & - & - & - \\
MVSNeRF$^c$~\cite{chen2021mvsnerf} + Semantic head & 53.8 & 62.9 & - & 55.3 & 69.7 & - & - & - & - \\
NeuRay$^c$~\cite{liu2022neural} + Semantic head & \underline{63.7} & 70.1 & - & \underline{77.5} & \underline{81.0} & - & - & - & - \\
Semantic Ray$^c$~\cite{liu2023semantic} & \textbf{76.0} & \textbf{80.8} & - & \textbf{91.1} & \textbf{94.0} & - & - & - & - \\
Semantic-NeRF$^d$~\cite{zhi2021place} & 18.7 & 46.1 & - & 15.4 & 31.3 & - & - & - & - \\
DINO-2D$^d$~\cite{caron2021emerging} & 18.1 & 46.1 & - & 20.6 & 35.5 & -  & - & - & - \\
DINO-NeRF$^d$~\cite{kobayashi2022decomposing} & 25.3 & 52.7 & - & 19.1 & 35.7 & - & - & - & - \\
JacobiNeRF 2D$^d$~\cite{xu2023jacobinerf} & 26.3 & 48.9 & - & 23.2 & 43.7 & - & - & - & - \\
JacobiNeRF 3D$^d$~\cite{xu2023jacobinerf} & 28.3 & 52.4 & - & 33.2 & 52.5 & - & - & - & - \\
Semantic-NeRF$^e$~\cite{zhi2021place} & 52.3 & \underline{72.8} & - & 42.1 & 61.9 & - & - & - & - \\
DINO-2D$^e$~\cite{caron2021emerging} & 33.5 & 62.4 & - & 34.4 & 52.5 & -  & - & - & - \\
DINO-NeRF$^e$~\cite{kobayashi2022decomposing} & 40.3 & 65.4 & - & 35.3 & 54.1 & - & - & - & - \\
JacobiNeRF 2D$^e$~\cite{xu2023jacobinerf} & 44.6 & 61.9 & - & 35.3 & 52.6 & - & - & - & - \\
JacobiNeRF 3D$^e$~\cite{xu2023jacobinerf} & 52.4 & 68.9 & - & 42.1 & 55.8 & - & - & - & - \\
\bottomrule
\end{tabular}
\vspace{-2ex}
\begin{flushleft}
\footnotesize{
$^a$The version with Slow-Fast contrastive loss; $^b$Setting of generalization; $^c$Setting of finetuning.\\
$^d$Sparse setting for segmentation propagation; $^e$Dense setting for segmentation propagation.
}
\end{flushleft}
\label{tab:room-seg}
\vspace{-6ex}
\end{table*}

Panoptic Quality (\textbf{PQ})~\cite{kirillov2019panoptic} is used in several works~\cite{fu2022panoptic, kundu2022panoptic} to evaluate the performance of panoptic segmentation. Specifically,
\begin{equation}
    PQ = \frac{\sum_{(p,g) \in TP} IoU(p,g)}{|TP| + \frac{1}{2} |FP| + \frac{1}{2} |FN|}.
\end{equation}
However, PQ does not measure whether instance identities are consistently preserved across views. To tackle this issue, a scene-level PQ is proposed in \cite{siddiqui2023panoptic}, denoted as $\textbf{PQ}^{\textbf{scene}}$, which modifies the matching process by considering the subsets that contain all segments belonging to the same instance or stuff class for a certain scene. There are several other metrics used for the evaluation of segmentation performance, such as Peak signal to noise ratio (\textbf{PSNR}), and mean of average precision (\textbf{mAP}). The average precision refers to the precision of an object detector for a specific class over different recall levels. For a particular class, the AP is calculated by plotting a Precision-Recall (P-R) curve, which plots the precision value (y-axis) against the recall value (x-axis) at different threshold levels. Then, the area under the curve (AUC) is computed, which gives the AP for that class.

\textbf{Results Summary}
We have summarized the quantitative results of segmentation from different methods based on the scale of scenes in
Table \ref{tab:room-seg} and Table \ref{tab:outdoor-seg}, respectively.

%% file: sections/6_pose_estimation.tex

\begin{figure}[t]
    \centering
    \includegraphics[width=\linewidth]{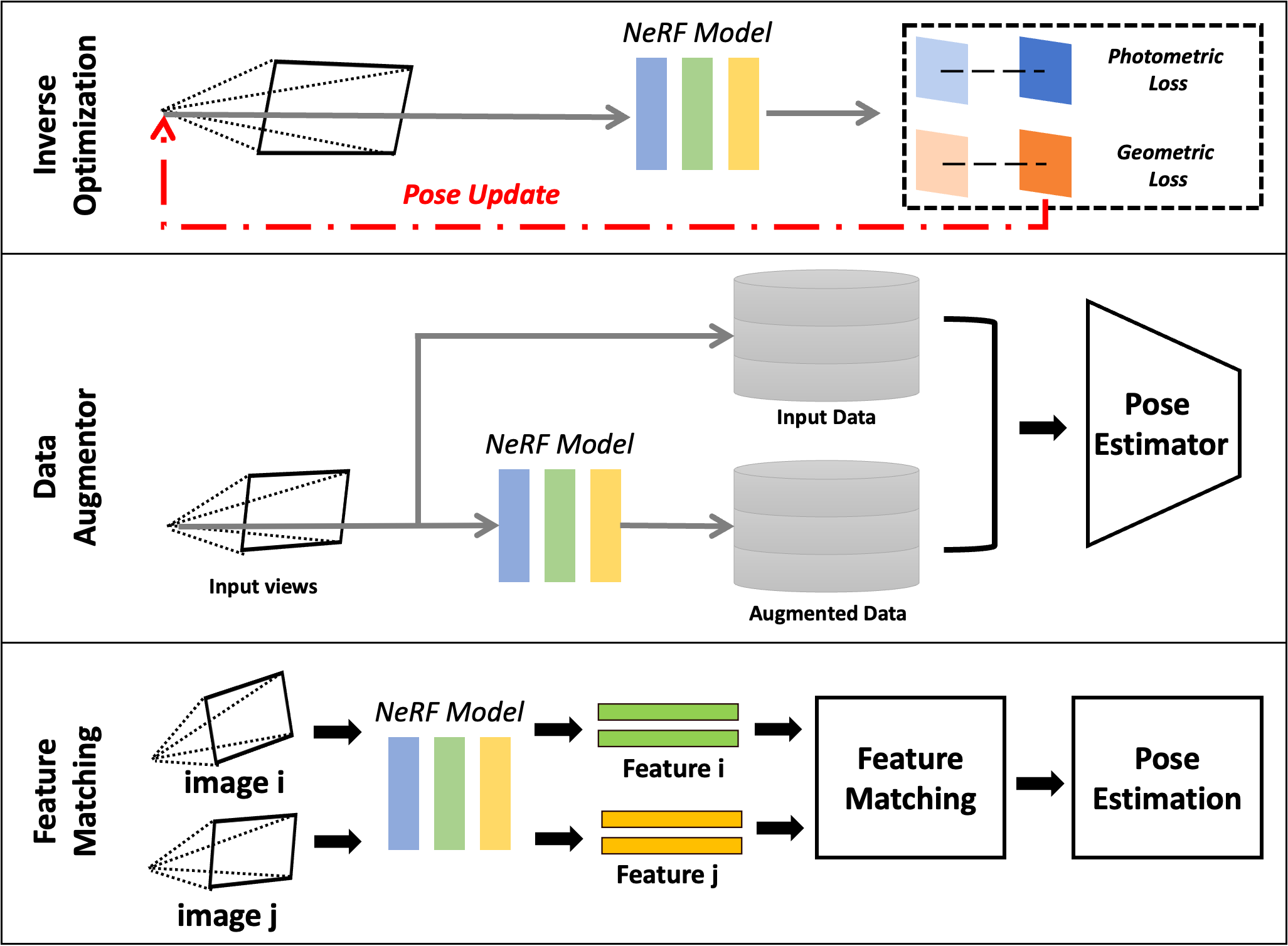}
    \caption{An overview of NeRF-based pose estimation pipeline.}
    \label{fig:pose}
\vspace{-4ex}
\end{figure}

Pose estimation, also referred to as localization, is a fundamental task in autonomous robots. It enables robots to ascertain their precise position and orientation relative to the surrounding environment. This capability underpins the entirety of robotic functionality, empowering robots to navigate complex terrains, avoid obstacles, manipulate objects with precision, and collaborate seamlessly with humans.
Consequently, as neural implicit fields like NeRF~\cite{nerf-c, nerf-j} have become a popular choice for 3D reconstruction tasks, many studies have delved into methodologies for executing pose estimation in a 3D scene represented by neural implicit fields.
Depending on how neural implicit fields are used in the pose estimation, we classify existing methods into the following three categories: 
\textit{inverse optimization}, \textit{data augmentor}, and \textit{feature matching}. At the end of this section, we also summarize the datasets and metrics, and evaluation results in \textit{Datasets and Evaluations}. An overview of the NeRF-based pose estimation pipeline is shown in Figure~\ref{fig:pose}.

\subsection{Inverse Optimization}
Given a 3D scene represented by neural implicit fields, the most straightforward way to estimate camera poses is to invert the NeRF optimization process. Specifically, unlike the optimization process of NeRF which fixes the poses of input views and updates the weights of the neural network, the pose estimation process fixes the weights of the neural network and updates the pose of the input views. Such an idea is firstly proposed by \textbf{iNeRF}~\cite{inerf} with the assumption that the intrinsic parameters of the camera are known. 
Concurrently, \textbf{NeRF$--$}~\cite{nerfmm} came to a similar idea of estimating camera poses. Though NeRF$--$ simultaneously optimizes the camera pose and the intrinsic parameters.
Although iNeRF and NeRF$--$ successfully performed pose estimation within a 3D scene represented by NeRF, they suffered from the low efficiency and the requirements of pre-training the network for the scene.

To address the former challenge,  \textbf{Lin \textit{et al.}}~\cite{lin2023icra} took advantage of the recent advances in 3D reconstruction and favored instantNGP~\cite{instantngp} instead of the vanilla NeRF~\cite{nerf-c, nerf-j}. To speed up the convergence, Lin \textit{et al.}~decomposed the $SE(3)$ camera pose into $SO(3)$ and $T(3)$ with first and second moments independently applied during the optimization. Finally, parallelized Monte Carlo sampling is proposed to enable optimization from multiple hypotheses. 
Further enhanced on this, \textbf{BID-NeRF}~\cite{bidnerf} is proposed with three main modifications. It first extended the objective function with a depth-based loss function, thus accelerating the convergence. Additionally, rather than using only a single input view, BID-NeRF sampled pixels from multiple view directions based on a moving window of frames and their relative transformations. Lastly, BID-NeRF discarded the time-consuming hierarchical sampling and only used the coarse model to estimate the camera pose.

Focusing on the second challenge, \textbf{SC-NeRF}~\cite{scnerf} took a sequential learning approach by first focusing on the scene geometry and the camera model, then optimizing for camera poses. 
\textbf{BARF}~\cite{barf}, on the other hand, is the first work to perform bundle adjustment on camera poses and the 3D scene represented by the neural implicit fields. It achieves bundle adjusting by simply setting both the weights of the neural network and the poses of the input views optimizable during the optimization. Such a design makes the model struggle to perform robustly in large-scale scenarios without a good initial pose estimation. 
Improved on this, \textbf{PoRF}~\cite{porf} introduced a pose residual field that regresses pose updates with a multi-layer perception (MLP). PoRF achieved better robustness with parameter sharing to leverage global information across the full sequence and an extra epipolar geometry loss to supervise the training without additional computation overhead. Similarly, \textbf{SPARF}~\cite{sparf} also enforced multi-view correspondence constraints with explicitly extracted feature points, but with a special focus on improving pose estimation performance on sparse input views. To the contrary, \textbf{L2G-NeRF} adopts a dense multi-view consistency by aligning multiple sets of 3D point clouds.
The idea of utilizing multi-view constraints can also be seen in \textbf{NoPe-NeRF}~\cite{nopenerf} and \textbf{NoPose-NeuS}~\cite{noposeneus}, which employ monocular depth estimation modules for extra depth supervision. 
On the other hand, \textbf{CBARF}~\cite{cbarf} proposed a coarse-to-fine pose estimation approach and boosted camera pose estimation performance with multiple bundle adjustment modules cascaded at different scales. In addition, CBARF introduced an erroneous pose detection module with hand-craft feature points to identify inaccurate poses and performed neighbor replacement to eliminate the detrimental effect of the inaccurately estimated camera pose.

Apart from estimating the pose of certain cameras, the 6 degrees-of-freedom (DoF) transformation between two neural radiance fields can also be estimated with inverse optimization. 
\textbf{nerf2nerf}~\cite{nerf2nerf} is one of the first works to investigate such problems. It relies on human-annotated keypoints to guide the optimization and gradually shifts from keypoints registration to NeRF registration throughout the optimization procedure. Improved on this, \textbf{Reg-NF}~\cite{regnf} presents an automated initialization with recognized objects in the scene. Specifically, it extracts and matches handcraft feature points on the objects and acquires an initial 6-DoF pose as a starting point for the inverse optimization.

\subsection{Data Augmentor}
Apart from directly inverting the optimization process, the ability of novel view synthesis is also exploited for pose estimation. \textbf{LENS}~\cite{lens}, as the pioneer, demonstrated better performance of pose regression networks when trained with extra synthetic data rendered by neural implicit fields. It first trained the NeRF-W~\cite{nerf-w} model on an outdoor scene and then used the real images and the rendered images to train CoordiNet~\cite{coordinet} for pose regression. 
Also using neural implicit fields to generate views at a given pose, \textbf{LATITUDE}~\cite{latitude} presented a two-stage localization approach that uses NeRF for both data augmentor and inverting optimization. In the first stage, LATITUDE adopted a trained Mega-NeRF~\cite{meganerf} to synthesize more views around the input image and train the pose regressor with both real and synthetic data. Then, in the second stage, LATITUDE further optimizes the regressed pose with the iNeRF-style optimization. A truncated dynamic low-pass filter is applied to the positional encoding to realize the coarse-to-fine optimization.

To improve the performance of pose regression networks, \textbf{DFNet}~\cite{dfnet} proposed to close the feature-level domain gap between the real and synthetic data. 
Its pipeline consists of a histogram-assisted NeRF, a feature extractor, and a pose regressor. With the histogram-assisted NeRF and the feature extractor separately trained, DFNet improves the pose regressor by matching feature maps extracted from the input image and the image rendered by the histogram-assisted NeRF.
Instead of performing pose regression on the appearance level, 
\textbf{NeRF-IBVS}~\cite{nerfibvs} proposes to perform 3D coordinate regression followed by image-based visual servo (IBVS) for pose refinement. It uses NeRF~\cite{nerf-c} as a data augmentor in two aspects. When training the coordinate regression network, it uses NeRF to provide estimated depth maps as pseudo-3D labels. Then in IBVS optimization, it uses NeRF to initialize the Jacobian matrix. 
Alternatively, pose regressor can be completely discarded and neural implicit fields can also serve as the data augmentor to densify the sparse map. Such an idea was first proposed in \textbf{IMA}~\cite{ima}, which is a hybrid approach that utilizes MLP-based implicit scene representation and 2D-3D feature matching for visual relocalization. In particular, IMA samples more accurate 3D points in the weak regions using the neural implicit representation, leading to more 2D-3D matches and smaller localization errors.

Moreover, NeRF can also be integrated with a particle-filter-based Monte Carlo localization~\cite{montecarloloc} approach for efficient pose estimation. Concentrating on 2D localization, \textbf{IR-MCL}~\cite{irmcl} presented a neural occupancy field to predict the occupancy probability of any given 2D location. Given this capability, novel LiDAR scans can be synthesized for arbitrary 2D poses. Thus, the trained neural occupancy field can be used as the observation model in the Monte Carlo localization framework. 
Similar ideas can also be found in \textbf{Loc-NeRF}~\cite{locnerf} for  3D localization tasks. Loc-NeRF continuously sampled candidate poses under the initial pose and used a trained NeRF~\cite{nerf-c} model to render novel views to find the correct pose direction.

\subsection{Feature Matching}
Lastly, as latent code has been widely used to encode local geometry in representing 3D scenes with neural implicit fields, such latent code can also be used for direct matching to estimate poses. Such an idea can be found in \textbf{Nerfels}~\cite{nerfels}, a locally dense yet globally sparse 3D representation. Nerfels defined latent code as continuous radiance fields within a small 3D sphere. Then, by formulating the Perspective-n-points (PnP) problem~\cite{epnp}, 3D poses can be estimated by minimizing the re-projection error and the photometric error. From a similar perspective, \textbf{FQN}~\cite{featurequerynetwork} attempted to 
balance between the accuracy of the scene geometry and the complexity of modeling high-dimensional scene descriptors. It focused on utilizing an MLP to model feature descriptors at given 3D locations from different viewing angles. The output descriptors of the MLP can then be used for pose estimation using either PnP formulation or direct alignment.

Also using 3D latent code, \textbf{NeuMap}~\cite{neumap} performs pose estimation by solving the proxy task of scene coordinate regression with the latent code that encodes a scene. Especially, a stack of scene-agnostic transformers is used to decode the coordinates and the confidence of key points detected with a pre-trained R2D2~\cite{r2d2} model. Using the points with enough confidence, the camera pose can be solved with PnP combined with RANSAC~\cite{ransac}. 
Similar ideas can also be found in the method proposed by \textbf{Altillawi \textit{et al.}}~\cite{altillawi2024ral}. Such a method regressed two sets of coordinates and predicted a set of confidence weights given any input image. The two regressed sets of coordinates are in the world coordinate frame and the camera coordinate frame respectively. Then, using the regressed point cloud and the predicted confidence weights, the pose of the input image can be estimated using dense alignment.
To obtain denser feature descriptors, \textbf{CROSSFIRE}~\cite{crossfire} proposed to learn an extra descriptor field to match the feature map extracted from the CNN and the corresponding one from the neural renderer. Then, using the depth map from the neural renderer, 3D feature points can be obtained and 2D-3D correspondences can be established. Thus, the camera pose of any input images can be solved with the PnP-RANSAC combination. 

Note that feature matching methods can also be appended with an inverse optimization step for better performance. For example, \textbf{PNeRFLoc}~\cite{pnerfloc} uses a scene-agnostic R2D2~\cite{r2d2} model to extract feature maps and then performs scene-specific feature adaption with an MLP. Then, the adapted features are used to train the PointNeRF~\cite{pointnerf} model and used to find 2D-3D correspondences for solving poses with the PnP-RANSAC combination. Finally, the poses are further optimized with iNeRF-style optimization using a warping loss. From another perspective, \textbf{NeRFMatch}~\cite{nerfmatch} directly learns the match between intermediate features in NeRF~\cite{nerf-c} and the image features extracted with a CNN, and estimates the camera pose with such 2D-to-3D correspondences. Then, using the correspondence loss along with the reconstruction loss, iNeRF-style optimization is performed for pose refinement.

\begin{table*}
\caption{Pose estimation accuracy in terms of median translation/rotation errors [meters/degrees]$\downarrow$ on 7-Scenes~\cite{7scenes} and Cambridge Landmarks~\cite{posenet}.}
\label{tab:loc-large-scale}
\footnotesize
\begin{tabular}{l|ccccccc|c}
  \toprule
  Scenes & Chess & Fire & Heads & Office & Pumpkin & Kitchen & Stairs & Avg. \\
  \midrule
  LENS\cite{lens} 
  & 0.03/1.3 & 0.10/3.7 & 0.07/5.8 & 0.07/1.9 & 0.08/2.2 & 0.09/2.2 & 0.14/3.6 & 0.08/3.0 \\
  DFNet$^a$\cite{dfnet} 
  & 0.04/1.5 & 0.04/2.2 & 0.03/1.8 & 0.07/2.0 & 0.09/2.3 & 0.09/2.4 & 0.14/3.3 & 0.07/2.2 \\ 
  NeRF-IBVS\cite{nerfibvs}
  & 0.03/1.2 & 0.03/1.2 & \underline{0.02}/1.1 & \underline{0.04}/1.4 & 0.08/2.4 & 0.07/2.2 & 0.08/2.1 & 0.05/1.6 \\ 
  IMA\cite{ima} 
  & \underline{0.02}/0.8 & \underline{0.02}/\underline{0.9} & \textbf{0.01}/\underline{0.9} & \textbf{0.03}/\underline{0.9} & 0.05/1.3 & 0.04/1.3 & 0.08/1.7 & 0.04/\underline{1.1} \\
  FQN$^b$\cite{featurequerynetwork} 
  & 0.04/1.3 & 0.05/1.8 & 0.04/2.4 & 0.10/3.0 & 0.09/2.5 & 0.16/4.4 & 1.40/34.7 & 0.27/7.2 \\
  NeuMap\cite{neumap} 
  & \underline{0.02}/0.8 & 0.03/1.1 & \underline{0.02}/1.2 & \textbf{0.03}/1.0 & \underline{0.04}/1.1 & 0.04/1.3 & \textbf{0.04}/\textbf{1.1} & \underline{0.03}/\underline{1.1} \\
  Altillawi \textit{et al.}$^c$ \cite{altillawi2024ral}
  & 0.05/1.4 & 0.16/4.6 & 0.09/6.1 & 0.13/3.5 & 0.12/2.5 & 0.10/2.5 & 0.17/3.5 & 0.12/3.4 \\
  CROSSFIRE\cite{crossfire} 
  & \textbf{0.01}/\underline{0.4} & 0.05/1.9 & 0.03/2.3 & 0.05/1.6 & \textbf{0.03}/\underline{0.8} & \underline{0.02}/\underline{0.8} & 0.12/1.9 & 0.04/\underline{1.1} \\
  PNeRFLoc\cite{pnerfloc} 
  & \underline{0.02}/0.8 & \underline{0.02}/\underline{0.9} & \textbf{0.01}/\textbf{0.8} & \textbf{0.03}/1.1 & 0.06/1.5 & 0.05/1.5 & 0.32/5.7 & 0.07/1.8 \\
  NeRFMatch\cite{nerfmatch}
  & \textbf{0.01}/\textbf{0.3} & \textbf{0.01}/\textbf{0.4} & \underline{0.02}/1.0 & \textbf{0.03}/\textbf{0.7} & \textbf{0.03}/\textbf{0.6} & \textbf{0.01}/\textbf{0.3} & \underline{0.07}/\underline{1.3} & \textbf{0.02}/\textbf{0.7} \\
  \midrule
  Scenes & St Mary’s & Court & Hospital & King’s & Shop Facade &  &  & Avg. \\
  \midrule
  LENS\cite{lens} 
  & 0.53/1.6 & -/- & 0.44/0.9 & 0.33/0.5 & 0.27/1.6 &  &  & 0.39/1.2 \\
  DFNet$^a$\cite{dfnet} 
  & 0.50/1.5 & -/- & 0.46/0.9 & 0.43/0.9 & 0.16/0.6 &  &  & 0.39/1.0 \\ 
  FQN$^b$\cite{featurequerynetwork} 
  & 0.58/2.0 & 42.53/0.8 & 0.54/0.8 & 0.28/0.4 & \underline{0.13}/0.6 &  &  & 0.38/1.0 (ex. Court) \\  
  NeuMap\cite{neumap} 
  & \underline{0.17}/\underline{0.5} & \textbf{0.06}/\textbf{0.1} & \underline{0.19}/\textbf{0.4} & \underline{0.14}/\textbf{0.2} & \textbf{0.06}/\textbf{0.3} &  &  & \underline{0.14}/\textbf{0.3} (ex. Court) \\
  Altillawi \textit{et al.}$^c$ \cite{altillawi2024ral}
  & 0.87/2.9 & -/- & 0.61/1.1 & 0.46/0.8 & 0.44/1.7 &  &  & 0.60/1.6 \\
  CROSSFIRE\cite{crossfire} 
  & 0.39/1.4 & -/- & 0.43/\underline{0.7} & 0.47/0.7 & 0.20/1.2 &  &  & 0.37/1.0 \\
  PNeRFLoc\cite{pnerfloc} 
  & 0.40/0.6 & 0.81/\underline{0.3} & 0.28/\textbf{0.4} & 0.24/\underline{0.3} & \textbf{0.06}/\textbf{0.3} &  &  & 0.25/\underline{0.4} (ex. Court) \\
  NeRFMatch\cite{nerfmatch}
  & \textbf{0.12}/\textbf{0.4} & \underline{0.23}/\textbf{0.1} & \textbf{0.09}/\textbf{0.4} & \textbf{0.06}/\textbf{0.2} & \textbf{0.06}/\underline{0.4} &  &  & \textbf{0.08}/\underline{0.4} (ex. Court) \\
  \bottomrule
\end{tabular}
\vspace{-2ex}
\begin{flushleft}
\footnotesize{
$^a$DFNet results are obtained with extra finetuning on validation images without the ground truth poses.\\
$^b$FQN results are obtained with MobileNetV2~\cite{mobilenetv2} backbone and K=30;\\
$^c$Altillawi \textit{et al.} results are obtained with MobileNetV3~\cite{mobilenetv3} backbone.
}
\end{flushleft}
\vspace{-4ex}
\end{table*}

\subsection{Datasets and Evaluations}

\textbf{Datasets:} Since inverse-optimization-based pose estimation methods are direct extensions of NeRF, datasets for the novel view synthesis task are widely adopted for evaluations. \textbf{LLFF}~\cite{llff}, as the most popular choice, is a dataset with 24 real-world scenes covering indoor and outdoor environments. Each scene contains 20-30 images taken with handheld cellphones, and the pseudo ground truth pose of these images is acquired with COLMAP~\cite{colmap-sfm}. In addition to LLFF, other datasets like \textit{Synthetic Objects}\cite{nerf-c, nerf-j}, \textit{BLEFF}~\cite{nerfmm}, \textit{MobileBrick}~\cite{mobilebrick}, \textit{DTU}~\cite{dtumvs}, \textit{BlendedMVS}~\cite{blendedmvs} are also evaluated in various inverse optimization methods. Moving on to more general pose estimation methods, 7-Scenes~\cite{7scenes} and Cambridge Landmarks~\cite{posenet} are preferred for evaluation. \textbf{7-Scenes} is a dataset with a collection of RGB-D sequences covering 7 indoor scenes. The pseudo ground truth camera poses and 3D reconstruction is acquired with the seminal KinectFusion~\cite{kinectfusion} system. \textbf{Cambridge Landmarks}~\cite{posenet}, on the other hand, is an outdoor urban dataset with 5 scenes. The dataset provides high-definition images subsampled from videos taken by smartphones. The scene and poses recovered with a structure from motion method~\cite{wu20133dv} is used as ground truth. Apart from these two, \textit{UMAD}~\cite{latitude}, \textit{Mill 19}~\cite{meganerf}, and \textit{Aachen Day \& Night} are used to evaluate city-scale pose estimation, \textit{NAVER Lab}~\cite{naver}, \textit{12-Scenes}~\cite{12scenes}, \textit{ScanNet}~\cite{scannet}, \textit{Replica}~\cite{replica} are selected to evaluate indoor pose estimation.

\textbf{Evaluation Metrics:} The most basic evaluation metrics for evaluating pose estimation are the translation error $\epsilon_{\bm{t}}$ and the rotation error $\epsilon_{\bm{R}}$. Based on these, more statistical evaluations can be carried out like median translation/rotation errors, root mean square errors (RMSE), accuracy given certain thresholds, \textit{etc}. Assume the estimated and the ground truth poses are:
\begin{equation}
    \bm{T}_{est} = 
    \begin{bmatrix}
        \bm{R}_{est} & \bm{t}_{est} \\
        \bm{0}^T & 1
    \end{bmatrix}, \;\;\; \bm{T}_{gt} = 
    \begin{bmatrix}
        \bm{R}_{gt} & \bm{t}_{gt} \\
        \bm{0}^T & 1
    \end{bmatrix}
\end{equation}
the translation and rotation errors can be computed with:
\begin{equation}
    \epsilon_{\bm{t}} = ||\bm{t}_{est} - \bm{t}_{gt}||_2
\end{equation}
\begin{equation}
    \epsilon_{\bm{R}} = \text{arccos}\frac{\text{tr}(\Delta \bm{R}) - 1}{2}
\end{equation}
where tr$(\cdot)$ is the trace operation and $\Delta \bm{R}=\bm{R}_{est}^T\bm{R}_{gt}$

\textbf{Results Summary:} Although many methods reviewed above-reported results on the LLFF~\cite{llff} dataset, different metrics like average pose error and accuracy given a threshold are used, making it very difficult to summarize them in a single table. Hence, we only summarize the results on 7-Scenes~\cite{7scenes} and Cambridge Landmarks~\cite{posenet} datasets in Table~\ref{tab:loc-large-scale}. Note that since many methods evaluated on the Cambridge Landmarks dataset didn't report their performance on the ``Trinity Great Court" scene, we only report the average performance on ``St. Mary Church", ``Shop Facade", ``King's College", and ``Old Hospital".

%% file: sections/8_SLAM.tex

\begin{figure*}[t]
    \centering
    \includegraphics[width=0.9\textwidth]{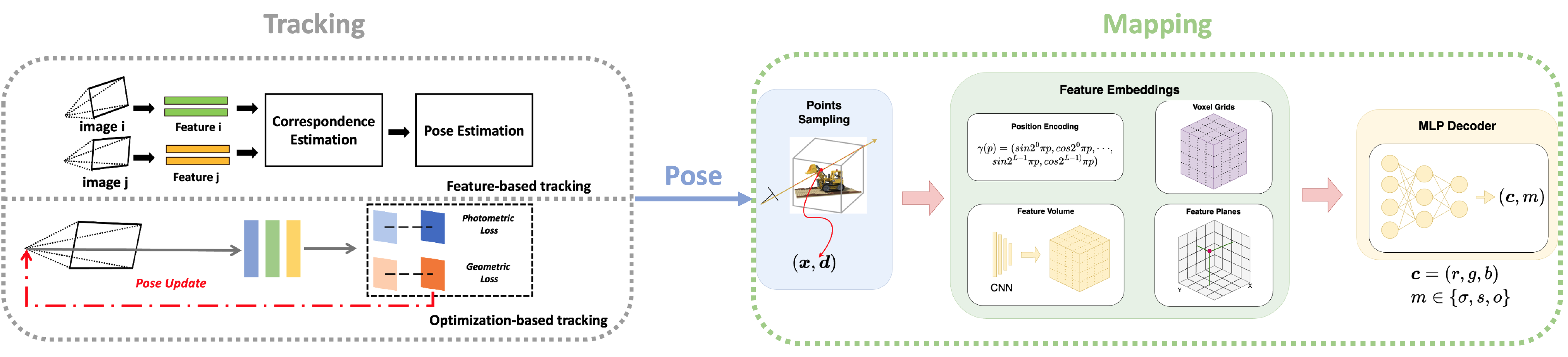}
    \caption{Overview of NeRF-based SLAM pipeline.}
    \label{fig:nerf-slam}
\vspace{-4ex}
\end{figure*}

SLAM is pivotal in the field of autonomous robots, aiming to construct maps of unfamiliar environments while determining the robot's position using onboard sensors. Perceptual devices categorize SLAM systems into visual SLAM and LiDAR SLAM.
Visual SLAM utilizes cameras, leveraging their affordability and rich environmental data for robust place recognition. LiDAR excels in pose estimation due to its accurate range measurements and resilience to environmental changes.
However, traditional SLAM faces challenges with data association, which refers to the task of matching sensor observations with real-world points, making geometry estimation challenging for unobserved regions. Implicit neural representations like NeRF overcome this limitation by synthesizing new views, making it promising for the SLAM system. In this section, 
we first review the methods
for NeRF-based SLAM systems in the order of \textit{RGB-D SLAM}, \textit{monocular SLAM}, and \textit{LiDAR SLAM}. Also, an overview of the NeRF-based SLAM pipeline is shown in Figure~\ref{fig:nerf-slam}.
Subsequently, we introduce the datasets and compare the results when evaluating NeRF-based SLAM methods.

\subsection{RGB-D SLAM}
Depth cannot be directly observed with a monocular camera, leaving the scale of the map and the trajectory estimation uncertain. As a result, visual SLAM systems commonly incorporate an additional depth sensor. Recently, NeRF-based RGB-D SLAM has drawn notable attention to integrating NeRF with SLAM for enhanced 3D reconstruction and camera tracking in dynamic environments. 

\textbf{iMAP}~\cite{iMAP} emerged as a pioneering work, merging RGB-D SLAM with NeRF to enable real-time 3D mapping and camera tracking in an end-to-end manner. iMAP adeptly captures and reconstructs evolving scenes by incrementally optimizing a NeRF model to estimate density and color as geometry and appearance representation. After optimization of the geometry and appearance, it inversely optimizes the input pose of the NeRF model for updating the tracking results. However, iMAP's reliance on a single MLP for neural mapping introduces memory issues as the mapping scale increases, leading to a forgetting problem. Therefore, \textbf{NICE-SLAM}~\cite{NICESLAM}, inspired by iMAP, improves the original method by employing voxel grid features to model the scene and decoding these features into occupancies using pre-trained MLPs. However, NICE-SLAM's ability to generalize to novel scenes is constrained due to the fixed pre-trained MLPs. In addition, the model's cubic memory growth rate tends to use low-resolution voxel grids, resulting in the loss of fine geometric details.

Despite the aforementioned limitations, numerous studies on neural implicit RGB-D SLAM have emerged since 2021, spurred by the foundational advancements of iMAP and NICE-SLAM. These subsequent works offer novel insights and approaches to enhance neural implicit representations and localization, spanning two critical domains: tracking and mapping. These areas reflect the dual aspects of SLAM systems where significant innovations have been observed.

In the \textit{mapping domain}, various approaches have been devised to accurately represent the geometry and appearance of environments. These methodologies primarily employ implicit representations, transforming world coordinates $\bm{x}$ into color $\bm{c}$ and various value types $m$. This value $m$ includes different forms of density representation, such as volume density $\sigma$, occupancy $o$, and SDF/TSDF $s$. Each form of representation serves a unique purpose, offering varying levels of detail and computational efficiency suited to different mapping needs.

Similar to iMAP, \textbf{MeSLAM}~\cite{kruzhkov2022meslam} adopts density $\sigma$ estimation, akin to the original NeRF~\cite{nerf-c}. Particularly, MeSLAM addresses SLAM's memory consumption using neural field representations, which enhance scene detail and accuracy by segmenting map regions across multiple networks. Moreover, \textbf{NID-SLAM}~\cite{xu2024nid} deploys an object removal and background restoration approach in a dynamic SLAM framework while leveraging an iMAP-alike neural implicit representation. At last, following the architecture of {Plenoxel}~\cite{yu2021plenoxels}, \textbf{RGBD PRF}~\cite{teigen2024rgb} bypasses neural networks and achieves an explicit geometry representation. This method integrates depth data into explicit sparse voxel grids, promoting a direct, neural-free approach to interpolate density $\sigma$, thereby facilitating environment reconstruction.

Drawing influence from NICE-SLAM, several approaches have adopted a voxel grid decoder for geometry representation, primarily focusing on modeling occupancy. Among them, \textbf{MLM}~\cite{li2023end} proposes a dense 3D reconstruction system with a multi-resolution feature encoding structure to accurately estimate occupancy. This system enhances detail extraction and updates locally, overcoming the limitations of a single MLP and significantly improving 3D reconstruction quality. \textbf{DNS-SLAM}~\cite{li2023dns} advances these ideas by introducing a coarse-to-fine semantic SLAM system, which utilizes hierarchical semantic representation for a multi-level understanding of scene semantics, integrating collaboration between appearance, geometry, and semantic features. Further enhancing this concept, \textbf{PLGSLAM}~\cite{deng2023plgslam} employs a progressive scene representation, merging local feature encoding with global MLPs for scalable indoor scene mapping. This approach fits well within the hierarchical mapping category due to its integration of multi-level features. \textbf{vMAP}~\cite{kong2023vmap} also employs multi-resolution feature grids to encode structure geometry into occupancy, targeting novel object-level dense SLAM system design. Diverging from these methods, \textbf{CP-SLAM}~\cite{hu2023cp} introduces a neural point-based 3D scene representation using multiple agents, facilitating flexible and dynamic map adjustments post-loop closure. This approach ensures seamless collaborative SLAM and delivers state-of-the-art performance in localization and reconstruction. Lastly, \textbf{Point-SLAM}~\cite{Sandström2023ICCV} offers a novel approach for dense SLAM using neural point clouds from RGB-D inputs. It dynamically anchors features within a generated point cloud, optimizing tracking and mapping through RGB-D based re-rendering loss minimization. This method adjusts anchor point density based on input detail, enhancing efficiency in sparse regions while capturing fine details in denser areas.

While many studies focus on density for representing geometry, SDF emerges as another way to represent surface geometry and is extensively employed across various studies. \textbf{iSDF}~\cite{ortiz2022isdf}, as the pioneering work to leverage SDF-based neural implicit representations for efficient 3D mapping. It uses a continuous neural network to represent signed distance fields and enables space-efficient encoding of scene geometry with adaptive levels of detail. In \textbf{SE3-Trans}~\cite{yuan2022ral}, SDF is leveraged for neural implicit map transformations to support loop closure in SLAM frameworks. The proposed model overcomes the non-remappable limitation of neural implicit maps by allowing transformations directly on these maps, facilitating remapping and demonstrating high-quality surface reconstruction capabilities. \textbf{Vox-Fusion}~\cite{yang2022vox} proposes a hybrid SLAM system that blends voxel-based mapping with neural implicit networks for efficient, detailed environment reconstruction using SDF representation. \textbf{Uni-Fusion}~\cite{yuan2024uni} also deploys an SDF-based implicit surface mapping, which deploys high-dimensional features such as SLIP embeddings. Furthermore, Uni-Fusion encodes point clouds into a latent implicit map (LIM), achieving incremental reconstruction by fusing local LIMs frame-wise into a global LIM. This versatile methodology not only focuses on incremental reconstruction, but also on other tasks like 2D-3D transfer, and open-vocab scene understanding. Similar to NeuS~\cite{neus}, \textbf{NIDS SLAM}~\cite{haghighi2023neural} implements an implicit mapping to learn SDF, and does not need to feed view direction to color MLP. This method combines ORB-SLAM3 tracking thread and loop closing thread with neural fields-based mapping for memory-efficient, dense 3D geometry. \textbf{SNI-SLAM}~\cite{zhu2023snislam} deploys a hierarchical implicit representation for cross-attention-based feature fusion, this coarse-to-fine semantic modeling enables the semantic embedding, color representation, and SDF surface reconstruction.

Just as SDF is a popular method for modeling geometry, TSDF is equally prevalent and widely deployed in many approaches for capturing the geometry and appearance of environments. \textbf{MIPS-Fusion}~\cite{tang2023mips} adopts a hierarchical strategy through its multi-implicit-submap scheme, allowing scalable TSDF-based surface reconstruction with rich local details by incrementally allocating and learning neural submaps alongside the scanning trajectory. \textbf{ESLAM}~\cite{eslam} processes RGB-D frames to incrementally build a scene representation, using multi-scale feature planes and shallow decoders to convert features into TSDF and RGB values. As a pair work, \textbf{Co-SLAM}~\cite{wang2023coslam} introduces a hybrid neural RGB-D SLAM system that offers robust camera tracking and high-fidelity surface reconstruction using TSDF in real-time. \textbf{ADFP}~\cite{Hu2023LNI-ADFP} uses a hierarchical approach by modeling low and high-frequency surfaces using different resolution feature grids, incorporating an attention mechanism to manage depth fusion from the TSDF. Unlike iSDF~\cite{ortiz2022isdf}, \textbf{iDF-SLAM}~\cite{ming2022idf} combines a feature-based deep neural tracker with an implicit single MLP mapper using TSDF for geometry representation, enabling self-supervised scene-specific feature learning for camera tracking and competitive scene reconstruction performance.

Conversely, within the \textit{tracking domain}, the primary distinction among studies arises from whether the SLAM front end derived from end-to-end learning methods or from conventional visual odometry (VO). While most of the approaches favor joint learning way for pose estimation, a subset of research opts for using conventional methods to initialize camera poses. To be specific, 
\textbf{NIDS SLAM}~\cite{haghighi2023neural} utilizes ORB-SLAM3~\cite{9440682} for the tracking thread. Similarly, \textbf{NGEL-SLAM}~\cite{mao2023ngelslam} employs a conventional feature-based tracking with loop closure, and neural implicit representation for dense mapping. Moreover, \textbf{NEWTOWN}~\cite{matsuki2023newton} uses ORB-SLAM2~\cite{murORB2} for initial pose estimation, while \textbf{OP SLAM}~\cite{10229827} utilizes the IMU information, it implements the pseudo motion data for tracking and a NeRF model for mapping.


\subsection{Monocular SLAM}
\label{subsec::mono}
Visual SLAM that utilizes only a monocular camera benefits from its cost-effectiveness and compact sensor setup. Recently, there have been significant advancements in implicit reconstruction using these types of cameras. These advancements can be broadly categorized into two main types.

The first type involves end-to-end joint mapping and tracking. This approach achieves simultaneous optimization of implicit scene representations and camera poses, enabling seamless mapping and tracking.
A recent application of this concept in dense RGB SLAM is illustrated by 
\textbf{NICER-SLAM}~\cite{zhu2023nicer}, which draws inspiration from NICE-SLAM. NICER-SLAM utilizes neural implicit representations for both mapping and camera tracking. This methodology facilitates end-to-end joint mapping and tracking under a neural representation framework by introducing additional supervisions compared to NICE-SLAM. These additional supervision include depths, normals, and optical flows. However, it's crucial to acknowledge that this pipeline has not been fully optimized for real-time performance at this stage.

Also drawing inspiration from NICE-SLAM, \textbf{Dense RGB SLAM}~\cite{li2023dense} propose an approach that uses a multi-resolution feature grid to represent the scene. In contrast to NICER-SLAM, which utilizes an off-the-shelf monocular predictor for depth prediction and enforces depth consistency between the rendered expected depths and monocular depths, this method incorporates a multi-view stereo loss to enforce consistent depth and color predictions among closely positioned images to jointly update camera poses and the map.
However, a potential concern arises as the MLP decoder is fixed after the initialization stage, which might cause problems when the camera moves to a region significantly different from those in the initialization.

The second type of advancement in monocular visual SLAM involves decoupling the tracking and mapping functions. This approach utilizes separate visual odometry for camera tracking while exclusively employing the NeRF model for mapping. As tracking and mapping operate independently, the system can leverage specialized visual odometry methods tailored for accuracy in camera tracking. Ensuring a reliable initial estimate for the camera pose is crucial, as it helps prevent the system from converging to local minima during mapping. Consequently, these methods depend on mature and reliable visual or visual-inertial odometry systems to provide robust and precise initial pose estimates.

For example, \textbf{VIO NeRF}~\cite{10003959} utilizes Apple ARKit to obtain the motion priors as initial camera poses to fit the NeRF network and then jointly optimize 3D scene representations and camera poses.

Similarly, \textbf{Orbeez-SLAM}~\cite{OrbeezSLAM} and \textbf{iMODE}~\cite{10161538} utilize ORB-SLAM series framework~\cite{7219438, murORB2, 9440682} to establish the initial pose. This approach benefits from ORB-SLAM's robust and mature system for accurate pose estimation in varied environments. Additionally, \textbf{DN-SLAM}~\cite{10376402}, which also uses the ORB-SLAM framework, uniquely incorporates dynamic object filtering to enhance 3D reconstruction in environments with moving objects.

\textbf{NeRF-VO}~\cite{naumann2023nerf} adopts a learning-based method in sparse visual odometry~\cite{teed2023deep} to estimate pose tracking and incorporate dense depth and surface normal supervision in a joint optimization between NeRF scene representation and camera poses. \textbf{NeRF-SLAM}~\cite{rosinol2022nerf} also utilizes a learning-based visual odometry, Droid-SLAM~\cite{teed2021droid}
to estimate initial accurate poses and depth maps with associated uncertainty. The objective of NeRF-SLAM is to achieve a real-time estimation of both geometric and photometric maps of the scene by leveraging techniques such as probabilistic volumetric fusion and InstantNGP (hash-based hierarchical volumetric radiance fields)~\cite{mueller2022instant}. 
However, \textbf{NeRF-SLAM} only performs offline global bundle adjustment at the end of camera tracking. In certain challenging scenarios, it's difficult to eliminate drift errors solely through offline refinement, as it lacks online loop closure detection and global bundle adjustment. 
This limitation restricts its ability to achieve globally consistent 3D reconstruction. Therefore, in addition to employing Droid-SLAM as the frontend for tracking, \textbf{HI-SLAM}~\cite{10374214} introduces an SDF to enhance surface definition and incorporates loop closing to ensure global consistency and \textbf{GO-SLAM}~\cite{zhang2023goslam} introduce an efficient loop closing and online full bundle adjustment.

\subsection{LiDAR SLAM}


LiDAR SLAM is renowned for its superior accuracy and direct provision of depth information in pose estimation, offering resilience against environmental variations such as changes in illumination and weather conditions. These features make it a preferred choice for many SLAM systems. However, traditional LiDAR SLAM tends to prioritize tracking quality, which often leads to a coarser scene representation due to the inherently sparse LiDAR data and the absence of RGB information. To overcome these limitations, integrating LiDAR SLAM with NeRF has emerged as a promising solution. This integration leverages NeRF's ability to generate dense, photorealistic scenes from sparse data, thus enhancing the environmental representation where traditional LiDAR systems fall short.

\textbf{LONER}~\cite{loner2023} first introduces the use of a neural implicit scene representation in the LiDAR SLAM system. The tracking module processes incoming LiDAR scans and estimates odometry using point-to-plane ICP. The scene is represented as an MLP with hierarchical feature grid encoding. To ensure depth supervision and prevent the NeRF model from forgetting learned geometry and experiencing slow convergence when integrating new LiDAR scans, the author optimizes the MLP using a dynamic margin loss combined with depth and sky losses (where weights are set to zero for rays pointing at the sky). \textbf{EINRUL}~\cite{yan2023efficient} does not utilize depth supervision due to its investigation into the discrepancy between LiDAR frames and depth images. It finds that a LiDAR frame contains only 5\% of the depth data found in a depth image, indicating that depth rendering may not offer sufficient geometry information for supervision. Therefore, EINRUL opts to supervise 3D space directly, eliminating the need for volume rendering and ground truth labels. This approach enables the refinement of initial poses alongside the optimization of the implicit representation. 

Since LiDAR SLAM is mainly utilized in outdoor environments, whereas NeRFs often struggle with outdoor, unbounded scenes. Addressing incremental LiDAR inputs with NeRF within large-scale environments presents a challenge. In response, an octree is typically employed to recursively split the scene into leaf nodes, each containing basic scene units known as voxels, enabling incremental representation of large-scale outdoor environments. Consequently, \textbf{NeRF-LOAM}~\cite{deng2023nerfloam} employs octree-based voxels alongside with neural implicit embeddings. These embeddings are converted into a continuous SDF by a neural implicit decoder. The joint optimization among embeddings, decoders, and poses is carried out by minimizing SDF errors.
\textbf{NF-Atlas}~\cite{yu2023nfatlas} adopts the concept of dividing the large-scale scene into multiple submaps using multi-resolution neural feature grids. In this approach, neural feature grids serve as nodes of the pose graph, while the relative pose between feature grids acts as the edges of the pose graph. This method ensures both local rigidity and global elasticity of the entire neural feature grid. Locally, NF-Atlas deploys a sparse feature octree and a compact MLP to encode the SDF of these submaps, enabling an end-to-end probabilistic mapping using Maximum a Posteriori (MAP). Transitioning to CLONeR~\cite{carlson2023cloner}, this system introduces occupancy grids to efficiently bypass ray marching in areas devoid of objects within unbounded scenes. CLONeR utilizes two separate MLPs: one to learn occupancy from LiDAR ray data and another to model the scene's RGB characteristics from camera rays. This dual-MLP configuration helps decouple the occupancy and visual data processing, ensuring that visual artifacts or illumination variances in the RGB images do not compromise the accuracy of the geometric information, such as the depth rendered from LiDAR data.

Maintaining global map consistency is also vital for improving reconstruction accuracy in large-scale environments. In this regard, \textbf{NF-Atlas} adjusts the volume origins to align with the robot trajectory revised by the occurrence of loop closure. On the other hand, \textbf{PIN-SLAM}~\cite{pan2024pinslam} adopts a different approach, utilizing a Point-Based Implicit Neural Representation to achieve global map consistency.

Reconstructing high-accuracy maps with resource-constrained devices has also emerged as a key area of research. Shi \textit{et al.}~\cite{10238795} tackle the challenge of generating precise maps in memory-limited hardware environments. They initiate the process by allocating a spatial hash grid near geometric surfaces using aligned batch range data. Points along rays within the grid are then sampled simultaneously. Each sampled point is transformed into a feature vector using a hierarchical hash encoder and then encoded into a TSDF value by a compact MLP.

\subsection{Datasets and Evaluations}
\textbf{Datasets:} When evaluating the above-mentioned methods, a large number of datasets are used. Regarding NeRF-based visual SLAM methods, three principle datasets stand out because of their extensive use in this field: \textbf{Replica}\cite{replica}, \textbf{TUM}\cite{TUM}, and \textbf{ScanNet}\cite{scannet}. These three datasets are particularly favored for providing comprehensive RGB and Depth images, as well as the ground truth poses. More importantly, these datasets focus on the indoor environment, which not only facilitates a detailed evaluation of mapping and tracking accuracy but also serves as a perfect scenario for testing the ability of novel view synthesis. In particular, Replica and ScanNet datasets have been introduced in Section~\ref{sec::recon_eval}, while the TUM dataset includes a variety of indoor scenes such as offices, a technical lab, household environments, and a large lecture room. For each sequence, the dataset provides synchronized RGB and depth images, camera intrinsics, and ground truth poses. The ground truth data is obtained using a motion-capture system with highly accurate resolution in smaller rooms or a high-precision laser tracker in larger rooms.

Moreover, regarding the NeRF-based Lidar SLAM methods, several datasets are introduced such as \textbf{KITTI}\cite{geiger2012we}, \textbf{MaiCity}\cite{vizzo2021poisson}, \textbf{Newer College}\cite{ramezani2020newer}, as well as \textbf{Fusion Portable}\cite{jiao2022fusionportable}. These datasets are critical for advancing NeRF-based LiDAR SLAM technologies. They provide diverse and challenging real-world scenarios, ranging from urban landscapes in KITTI and MaiCity to academic campus settings in Newer College and varied portable scenarios in Fusion Portable. These datasets offer rich LiDAR data along with accurate ground truth for localization and mapping, enabling the evaluation of LiDAR-based SLAM systems under different conditions and scales. This variety aids in robust testing and refinement of NeRF techniques tailored to LiDAR data, highlighting their adaptability and effectiveness across various environments. However, it is important to note that this section does not provide metric comparisons for the aforementioned LiDAR datasets. This absence is due to the lack of a common metric benchmarking across the papers reviewed in our paper. Each study often utilizes different evaluation criteria, making direct comparison challenging. This highlights a gap in the field, underscoring the need for standardized metrics to better assess and compare the performance of NeRF-based LiDAR SLAM systems across diverse datasets.


\textbf{Evaluation Metrics:} We evaluate the performance of the aforementioned methods in terms of their mapping accuracy and tracking capabilities. First, We provide an overview of popular metrics, namely Accuracy (Acc.), Completion (Comp.), Completion Ratio (C.R. $<$ 5cm $\%$), and Depth (Dep. L1) when evaluating the mapping performance. 
With the first three already discussed in Section~\ref{sec::recon_eval}, we have Dep. L1 as:
\begin{equation}
\text { Dep.~L1 }=\frac{1}{\left|P_{p r e d}\right|} \sum_{p \in P_{p r e d}}\left|d_p-d_{q(p)}\right|
\label{equ:depthl1}
\end{equation}


Secondly, by following \cite{sturm2012benchmark}, we evaluate the tracking performance by computing root mean squared error (RMSE). Given $y_i$ as the estimated pose and $\hat{y}_i$ as the ground truth pose, we can compute RMSE:

\begin{equation}
\mathrm{RMSE}=\sqrt{\frac{1}{N} \sum_{i=1}^N\left(y_i-\hat{y}_i\right)^2}
\end{equation}
\begin{table}[t]
\caption{Mapping quality of room-scale environments on Replica~\cite{replica} datasets, the results without special marks are taken from their original papers.}
\footnotesize
\begin{tabular}{c|l|cccc}
\toprule
  \multicolumn{2}{c|}{Methods}  & Acc.{$\downarrow$} & Comp. {$\downarrow$} & C.R. {$\uparrow$} & Dep. L1 {$\downarrow$}\\
  \midrule
  \parbox[t]{2mm}{\multirow{20}{*}{\rotatebox[origin=c]{90}{RGB-D}}}
  & iMAP~\cite{iMAP} &  6.95  &  5.33 & 66.60 & 7.64\\
  & NICE-SLAM~\cite{NICESLAM} &   2.85  &  3.00 & 89.33 & 3.53\\
  & NID-SLAM~\cite{xu2024nid} &    2.72 & 2.56  & 91.16 & 2.87\\
  & RGBD PRF~\cite{teigen2024rgb} &    2.72 & 2.56  & 91.16 & 2.87\\
  & MLM~\cite{li2023end} & 3.64  &  2.80 & 90.43 & 2.30\\
  & DNS-SLAM~\cite{li2023dns} &  2.76  & 2.74  & 91.73 &3.16\\
  & PLGSLAM~\cite{deng2023plgslam} &  \underline{1.79}  & \underline{1.54}  & \underline{97.88} &0.77\\
  & vMAP~\cite{kong2023vmap} &   3.20  & 2.39  & 92.99 & -\\
  & CP-SLAM~\cite{hu2023cp} &  3.33  & -  & - &1.38\\
  & Point-SLAM~\cite{Sandström2023ICCV} & -& -&- & \textbf{0.44}\\

  & SE3-Trans~\cite{yuan2022ral}   &  2.17 & 5.72  &  80.66&-\\
  & Vox-Fusion~\cite{yang2022vox}   & 2.37  & 2.28  &  92.86&-\\
  & Uni-Fusion~\cite{yuan2024uni} &- &- & -&1.47 \\
  & NIDS SLAM~\cite{haghighi2023neural} & -& -& 67.3 & 0.56 \\
  & SNI-SLAM~\cite{zhu2023snislam} &1.94 & 1.70& 96.62& \underline{0.77}\\
  & MIPS-Fusion~\cite{tang2023mips} & 4.43 &  - & 69.7 & -\\
  & ESLAM~\cite{eslam}  & \textbf{0.97}  &  \textbf{1.05} & \textbf{98.60} & 1.18\\
  & Co-SLAM~\cite{wang2023coslam}    &  2.10 & 2.08  & 92.99 &1.51\\
  & ADFP~\cite{Hu2023LNI-ADFP} & 2.59 & 2.28 & 93.38 & 1.81\\
  & iDF-SLAM~\cite{ming2022idf} & 4.86 &  3.05 & 86.34 & -\\  
  \midrule
  \parbox[t]{2mm}{\multirow{9}{*}{\rotatebox[origin=c]{90}{RGB}}}
  & NICER-SLAM~\cite{zhu2023nicer}  & 3.65  & 4.16  & 79.37 &-\\
  & Li \textit{et al.}~\cite{li2023dense}    & 4.03  & 4.20  & 79.60 &- \\
  & VIO NeRF\cite{10003959} & -  & -  & - &-\\
  & Orbeez-SLAM\cite{OrbeezSLAM}   &  - &  - &  -& 11.88\\
  & iMODE~\cite{10161538}   & 8.78  &  13.9 &  37.10 &-\\
  & NeRF-VO~\cite{naumann2023nerf}   & \textbf{2.81}  &  \textbf{3.59} & 81.30 &-\\
  & NeRF-SLAM~\cite{rosinol2022nerf}  & \underline{3.10}  & 4.08  & \textbf{85.80} & 4.49\\
  & HI-SLAM\cite{10374214}   & 3.56  &  \underline{3.60} & \underline{82.95} & \textbf{3.63}\\
  & GO-SLAM\cite{zhang2023goslam} & 3.81 & 4.79 & 78.00 & \underline{4.39} \\
\bottomrule
\end{tabular}
\label{tab:slam-room-mapping}
\vspace{-4ex}
\end{table}

\begin{table*}
\caption{Tracking results of room-scale environments on Replica~\cite{replica} datasets. We mainly report the RMSE[cm] results in this table.}
\footnotesize
\begin{tabular}{c|l| ccccccccc }
\toprule
  \multicolumn{2}{c|}{Methods}  & room-0 & room-1 & room-2 & office-0 & office-1 & office-2 & office-3 & office-4 & Avg. \\
  \midrule
\multirow{15}{*}{RGB-D}  & {iMAP~\cite{iMAP}}  & 5.23  &  3.09 & 2.58 & 2.40 & 1.17 & 5.67 & 5.08 & 2.23& 3.42\\ 
  & {NICE-SLAM~\cite{NICESLAM}} &  1.69 & 2.04 & 1.55 & 0.99 & 0.90 & 1.39 & 3.97 & 3.08 & 1.95\\
& RGBD PRF~\cite{teigen2024rgb} & 0.93 & 0.61 & 1.79 & 1.17 & \textbf{0.27} & 0.56 & 0.72 & \underline{0.61} & 0.83 \\
 &DNS-SLAM~\cite{li2023dns} & {0.49} & 0.46 & \underline{0.38} & \textbf{0.34} & \underline{0.35} & \textbf{0.39} & {0.62} & \textbf{0.60} & \textbf{0.45} \\
 & PLGSLAM~\cite{deng2023plgslam} & -& -&- &- &- &- & -&- & 0.64 \\
& CP-SLAM$^*$~\cite{hu2023cp} & \underline{0.48} & \underline{0.44} & -& 0.56 & -& -& \textbf{0.37} &- & \underline{0.46} \\
 &Point-SLAM~\cite{Sandström2023ICCV} & 0.61 & \textbf{0.41} & \textbf{0.37} & \underline{0.38} & 0.48 & \underline{0.54} & 0.72 & 0.63 & 0.52 \\
 & Vox-Fusion~\cite{yang2022vox} & \textbf{0.40} & 0.54 & 0.54 & 0.50 & 0.46 & 0.75 & \underline{0.50} & \textbf{0.60} & 0.54 \\
 & NIDS-SLAM~\cite{haghighi2023neural}  & 0.58 & \textbf{0.41} & 0.58 & 0.62 & {0.40} & 1.20 & 0.88 & 1.80 & 0.80 \\
 & SNI-SLAM~\cite{zhu2023snislam} &- & -&- &- &- &- &- & -& \underline{0.46} \\
 & MIPS-Fusion~\cite{tang2023mips} & 1.10 & 1.20 & 1.10 & 0.70 & 0.80 & 1.30 & 2.20& 1.10  & 1.19\\
 &ESLAM~\cite{eslam} & 0.76 & 0.71 & 0.56 & 0.53 & 0.49 & 0.58 & 0.74  &0.64 & 0.62 \\
 &Co-SLAM~\cite{wang2023coslam} & 0.65 & 1.13 & 1.43 & 0.55 & 0.50 & 0.46 & 1.40 & 0.77 & 0.86 \\
 &ADFP~\cite{Hu2023LNI-ADFP} & 1.39 & 1.55 & 2.60 & 1.09 & 1.23 & 1.61 & 3.61 & 1.42 & 1.81 \\
 &iDF-SLAM~\cite{ming2022idf} & 1.80 & 5.90 & 2.60 & 1.60 & 2.10 & 1.80 & 1.90 & 2.10 & 2.50 \\
  \midrule
\multirow{6}{*}{RGB}  & {NICER-SLAM~\cite{zhu2023nicer}} &  1.36 & 1.60 & 1.14 & 2.12 & 3.23 & 2.12 & 1.42 & 2.01 & 1.88\\
& Li \textit{et al.}~\cite{li2023dense} & 0.48 & 0.78 & \underline{0.35} & 0.67 & \textbf{0.37} & \textbf{0.36} & \textbf{0.33} & \textbf{0.36} & 0.46 \\
& DDN-SLAM$^*$~\cite{li2024ddn} & - & - & - & 5.80 & 7.10 & 1.10 & 5.60 & 1.20  & 4.27\\
& NeRF-VO~\cite{naumann2023nerf} & \textbf{0.28}  &\textbf{0.33} & 0.36 & \underline{0.53} & \underline{0.40} & \underline{0.53} & \underline{0.56} & \underline{0.47} & \underline{0.43}\\
& NeRF-SLAM~\cite{rosinol2022nerf} &\underline{0.40} & \underline{0.61} & \textbf{0.20} & \textbf{0.21} & 0.45  &0.59  &\textbf{0.33}  &1.30 & 0.51 \\
& Go-SLAM~\cite{zhang2023goslam} & -& -&- &- & -& -& -&- & \textbf{0.39} \\
\bottomrule
\end{tabular}
\label{tab:slam-tracking-replica}
\vspace{-2ex}
\end{table*}

\begin{table*}
\caption{Tracking results of room-scale environments on TUM~\cite{TUM} and ScanNet~\cite{scannet} datasets. We mainly report the RMSE[cm] results in this table.}
\footnotesize
\begin{tabular}{c|l| cccc|ccccccc }
\toprule
\multicolumn{2}{c|}{ } & \multicolumn{4}{c|}{TUM} & \multicolumn{7}{c}{ScanNet} \\
  \multicolumn{2}{c|}{Methods}  & fr1/desk & fr2/xyz & fr3/office & Avg. & 0000 & 0059 & 0106 & 0169 & 0181 & 0207 & Avg. \\
  \midrule
\multirow{17}{*}{RGB-D}  & {iMAP~\cite{iMAP}} &  4.90 &  2.00 &  5.80 &  4.23 & 55.95 &  32.06 &  17.50 &  70.51 &  32.10 &  11.91 &  36.67  \\
  & {NICE-SLAM~\cite{NICESLAM}}& 2.70 & 1.80 & 3.00 & 2.50 & 8.64 & 12.25 & 8.09 & 10.28 & 12.93 & 5.59 & 9.63 \\
& MeSLAM~\cite{kruzhkov2022meslam} & 6.00 & 6.54 & 7.50 & 6.68 &- & -&- &- & -& -& -\\
& RGBD PRF$^*$~\cite{teigen2024rgb} &- & -&- & -& \textbf{2.46} & - & \textbf{3.87} & \textbf{1.65} & \textbf{3.73} & \textbf{2.59} & \textbf{2.86}  \\ 
 & {MLM}~\cite{li2023end} & \underline{2.40} & 1.70 & 2.90 & 2.33 & 6.90 & 9.10 & 7.40 & \underline{3.10} & \underline{8.60} & - & {7.00} \\
& DNS-SLAM~\cite{li2023dns} &- &- & -& -& \underline{5.42} & \textbf{5.20} & 9.11 & 7.70 & 10.12 & 4.91 & 7.07 \\
 & PLGSLAM~\cite{deng2023plgslam} &- & -&- & -& -&- & -& -&- & -& \underline{6.77} \\
& vMAP~\cite{kong2023vmap} & 2.60 & 1.60 & 3.00 & 2.40 &- & -&- &- & -& -& -\\
& Point-SLAM~\cite{Sandström2023ICCV} & - &-&- &- & 10.24 & 7.81 & 8.65 & 22.16 & 14.77 & 9.54 & 13.97 \\ 
& Vox-Fusion~\cite{yang2022vox} &- & -& -& -& 8.39 &- & 7.44 & 6.53 & 12.20 & 5.57  & 8.03\\
& Uni-Fusion~\cite{yuan2024uni} & \textbf{1.80} & \textbf{0.50} & \textbf{2.10} & \textbf{1.47} &- &- &- &- & -& -&- \\
& SNI-SLAM$^*$~\cite{zhu2023snislam} &- &- &- & -& 6.90  &\underline{7.38}  &7.19 & -&- & \underline{4.70}  &6.54\\
& MIPS-Fusion~\cite{tang2023mips} &3.00 & 1.40& 4.60&3.00 
& 7.90 & 10.70 & 9.70 & 9.70 & 14.20 & 7.80 & 10.00
\\
& ESLAM~\cite{eslam}  & 2.47 & \underline{1.11} & 2.42 & \underline{2.00} & 7.30 & 8.50 & 7.50 & 6.50 & 9.09 & 5.70 & 7.42 \\ 
& Co-SLAM~\cite{wang2023coslam}  & \underline{2.40} & 1.70 &  \underline{2.40}& 2.17& 7.18 & 12.29 & 9.57 & 6.62 & 13.43 & 7.13 & 9.37 \\
& ADFP~\cite{Hu2023LNI-ADFP} & -&- & -& -& - & 10.50 & 7.48 & 9.31 & - & 5.67 & 8.24 \\
 &iDF-SLAM~\cite{ming2022idf} &- & -& -&- & 57.70  &8.20  &\underline{5.80} & 39.90 & 29.10 & 15.40 &26.00 \\
  \midrule

\multirow{7}{*}{RGB}  & {NICER-SLAM~\cite{zhu2023nicer}} &  6.95  &  5.33 & 66.60 & 7.64 &- &- & -&- &- & -&- \\
& Li \textit{et al.}~\cite{li2023dense} & \underline{2.00}  & \underline{0.60} & 2.30 & 1.60 &- &- &- &- &- &- &-\\
& Orbeez-SLAM~\cite{OrbeezSLAM} & 1.90  & \textbf{0.30} & \textbf{1.00} & \textbf{1.07} & 7.22 & \textbf{7.15} & \underline{8.05}  &\textbf{6.58} & 15.77  &\textbf{7.16}  &8.65\\
& NeRF-VO~\cite{naumann2023nerf} & -& -&- &-&12.70 & 19.00 & 12.40 & 13.20 & 9.00 & 8.60 & 12.50 \\
& NeRF-SLAM~\cite{rosinol2022nerf} & -& -& -&- &14.90  &16.60  &10.70 & 16.50 & 12.80 & 13.80 & 14.20 \\
& HI-SLAM~\cite{10374214} &- &- &- & -& \underline{6.40} & \underline{7.20} & \textbf{6.50} & \underline{8.50} & \underline{7.60} & \underline{8.40} & \textbf{7.40} \\
& Go-SLAM~\cite{zhang2023goslam} & \textbf{1.50} & \underline{0.60} & \underline{1.30} & \underline{1.13} & \textbf{5.94} & 8.27 & 8.07  &8.42 & \textbf{8.29} & -& \underline{7.79} \\
\bottomrule
\end{tabular}
\label{tab:slam-tracking-tum-scannet}
\vspace{-4ex}
\end{table*}


\textbf{Results Summary:} Finally, we summarize the reported results as shown in Table~\ref{tab:slam-room-mapping}, Table~\ref{tab:slam-tracking-replica}, Table~\ref{tab:slam-tracking-tum-scannet}. Note that we use bold font to represent the best performance while use underline to demonstrate the second best method. Particularly, the mapping quality of room-scale environments on the Replica dataset is shown in Table~\ref{tab:slam-room-mapping}. Among the RGB-D methods, ESLAM emerged as the standout performer across all metrics in the Replica dataset, showcasing unparalleled mapping quality with the highest Completion Ratio (98.60\%) and lowest Accuracy (0.97 cm) and Completion (1.05 cm). PLGSLAM followed closely, delivering impressive Completion and Accuracy, underlining the effectiveness of depth data in achieving high-fidelity reconstructions. For RGB methods, NeRF-SLAM exhibited strong Completion and Completion Ratio, highlighting the potential of purely image-based approaches in achieving detailed environmental mappings. Moreover, tracking results on the Replica dataset among 8 environments are listed in Table~\ref{tab:slam-tracking-replica}, while tracking results on the TUM and ScanNet dataset are shown in Table~\ref{tab:slam-tracking-tum-scannet}. Specifically, in tracking accuracy on the Replica dataset, DNS-SLAM led the RGB-D category with an average RMSE of 0.45 cm, closely followed by CP-SLAM and Point-SLAM, highlighting strong localization and mapping accuracy. RGB techniques, with Li \textit{et al.} and NeRF-VO at the forefront, also showed promising tracking performance, indicating the potential of image-based SLAM to compete with depth-based methods.

For the TUM and ScanNet datasets, Uni-Fusion and Orbeez-SLAM outperformed other RGB-D methods in tracking accuracy, proving the value of depth data for precise localization across varied environments. Meanwhile, Go-SLAM and HI-SLAM excelled in the RGB category, reflecting progress in visual SLAM technologies for navigating complex indoor spaces.

%% file: sections/9_plan_navi.tex
NeRF can significantly enhance planning in autonomous robots by providing high-quality 3D representations of environments from 2D images. This technology enables robots to accurately perceive and navigate through complex environments, improving their path planning, obstacle avoidance, and object interaction capabilities. By leveraging NeRF's ability to create detailed and realistic 3D maps, robots can make more informed decisions, leading to safer and more efficient navigation and task execution in a variety of settings. The realistic mapping provided by NeRF is especially crucial in dynamic environments where real-time data interpretation and spatial awareness are paramount.
\subsection{Navigation}
\textbf{NeRF-Nav} \cite{adamkiewicz2022vision} proposes a navigation system based on a pre-trained NeRF. The robot's objective is to navigate through free space in NeRF to reach the goal pose. To achieve this, the authors introduce a trajectory optimization algorithm for avoiding collisions in the high-density areas of NeRF and an optimization-based filtering to estimate the pose and velocity of the equipped camera. The final navigation system is constructed by combining the trajectory optimization and filtering together in an online replanning loop. Instead of deriving the collision information from the density in NeRF, \textbf{CATNIPS} \cite{chen2023catnips} introduces a transformation of NeRF to a Poisson Point Process (PPP) \cite{kingman1992poisson}, which allows for rigorous uncertainty quantification of collision in NeRF. Then a voxel representation called Probabilistic Unsafe Robot Region (PURR) is introduced to consider the robot geometry in trajectory optimization.

Visual localization and navigation are important tasks in robotics --- given a query image, finding the location of this image in the environment, and navigating to reach the pose that has the same view as the query image. \textbf{RNR-Map} \cite{kwon2023renderable} addresses those challenges by leveraging a latent-code-based NeRF \cite{devries2021unconstrained}. Similar to conventional occupancy maps, the constructed map has a grid form but the information stored in each grid is not the occupancy but a latent code, which is used as an intermediate representation of the image at the corresponding place. The latent code is converted from the input image by an encoder and is decoded to an input to the NeRF model \cite{devries2021unconstrained} by a decoder. 
Different from mobile navigation in indoor environments \cite{kwon2023renderable}, the task of visual localization and navigation can also happen to manipulators, as in \textbf{NSR-VC}.
Specifically, a time contrastive loss is combined with NeRF in an autoencoding framework to learn viewpoint-invariant scene representations.
Such a representation is later used to enable visuomotor control for challenging manipulation tasks. Similar to the latent-code-based NeRF \cite{kwon2023renderable, devries2021unconstrained}, \textbf{Comp-NeRF} uses latent vectors to parameterize individual NeRFs from which the scene can be reconstructed. Then a graph neural network is trained in the latent space to capture compositionality in dynamic multi-object scenes. 

Another way of visual navigation is to use control barrier functions (CBFs) in feedback controllers. However, CBFs are poorly suited to visual controllers due to the need to predict visual features. To address this issue, \textbf{CBF-NeRF} \cite{tong2023enforcing} leverages NeRF to provide single-step visual foresight for a CBF-based controller. In this way, the controller is able to determine whether an action is safe or unsafe using only a single-step horizon. \textbf{MultiON} \cite{marza2023multi} proposes to use implicit representations --- a semantic representation and an occupancy representation in Reinforcement Learning. The two representations are modeled by two neural networks, whose weights can be updated during training.

\subsection{Active Mapping}
\textbf{UGP-NeRF} \cite{lee2022uncertainty} studies how a mobile robot with an arm-held camera can select a favorable number of views to recover an object’s 3D shape efficiently. To achieve the goal, the authors introduce a ray-based volumetric uncertainty estimator, which computes the entropy of the weight distribution of the color samples along each ray of the object’s implicit neural representation. Based on the guidance of the uncertainty estimation, a next-best-view selection policy is proposed to select the efficient views. Instead of using the entropy of weight distribution as the uncertainty quantification, \textbf{NeruAR} \cite{ran2023neurar} treats the prediction of color for each sampled point as a Gaussian distribution, whose mean and variance are optimized by the rendering loss. Then the variance is found to be connected with the metric of Peak Signal-to-Noise Ratio (PSNR) by a linear relationship. In this case, the variance can be treated as a proxy of PSNR for evaluating the image quality. Modeling the color as a distribution is also used in \textbf{ActiveNeRF}~\cite{activenerf}, where the authors propose to select the samples that bring the most information gain by evaluating the reduction of uncertainty given new inputs to enhance the training of NeRF. \textbf{AutoNeRF} \cite{marza2023autonerf} considers using an autonomous embodied agent to collect data for training a NeRF. To collect the data for NeRFs, a modular exploration policy \cite{chaplot2020object, chaplot2020learning} is trained using several reward signals, including explored area, obstacle coverage, semantic object coverage, and viewpoints coverage. After the policy training, the trained policy is used to collect data in unseen environments, and the collected data is further used for training the NeRF model. The resultant NeRF can be used in a series of downstream tasks, including reconstruction, planning, mapping, rendering, and pose refinement. \textbf{ActiveRMAP}~\cite{activermap} uses the DVGO \cite{sun2022direct} as the NeRF model and presents an RGB-only active vision framework for active 3D reconstruction and planning in an online manner. The planning is achieved by a multi-objective optimization formulation that takes several factors into account, including collision avoidance, information gain, and path efficiency. \textbf{NARUTO}~\cite{feng2024naruto} incorporates an uncertainty learning module that dynamically
quantifies reconstruction uncertainty while actively reconstructing the environment into a multi-resolution hash grid as the mapping backbone. A novel uncertainty aggregation strategy is proposed to leverage the learned uncertainty for goal searching and efficient path planning. 

\subsection{Datasets and Evaluations}
Datasets and metrics used for planning and navigation are highly varying across different works. In this section, we provide a descriptive summary of the ones used in the works reviewed by this survey.

\textbf{Datasets:}
Several public image-goal navigation datasets are involved. \textbf{NRNS}~\cite{hahn2021no} dataset contains
straight and curved paths, which are classified into easy, medium, and hard, depending on the navigation difficulty.
Apart from the hard-straight set, which only has 806 episodes, each difficulty-path type combination encloses 1000 episodes. \textbf{Gibson}~\cite{xia2018gibson} dataset, as another popular in-house dataset for navigation tasks, has 86 houses in total, with 72 for training and 14 for validation. 
In addition, \textbf{MP3D}~\cite{chang2017matterport3d} is a comprehensive RGB-D dataset. It covers 90 building-scale scenes and provides 10,800 panoramic views from 194,400 RGB-D images. It also provides annotations for various tasks, including 3D reconstructions, camera pose estimation, and 2D and 3D semantic segmentation. Among all the reviewed papers, some works, such as NSR-VC~\cite{li20223d}, leverage existing simulators, e.g., NVIDIA FleX~\cite{macklin2014unified} to generate environments for experiments. Other simulators used include PyBullet~\cite{coumans2021} and Habitat~\cite{savva2019habitat}.

\textbf{Evaluation Metrics:}
Two experimental goals can be observed from all reviewed works --- one is using NeRF as a perception model and aiming to achieve more efficient navigation behaviors; another one is to leverage the navigation capability of robots to obtain 3D models with higher quality. Metrics for the former task aim to evaluate the performance of navigation, such as Success Rate (SR) and Success weighted by Path Length. For the latter task, there are also several metrics to evaluate the 3D reconstruction results, such as F-score~\cite{lee2022uncertainty, chen2023planarnerf}, IoU, and Chamfer distance between the reconstructed model and the ground-truth model. Other used metrics include accuracy, completion, completion ratio, and mean absolute distance \cite{chen2023planarnerf, feng2024naruto}.

%% file: sections/10_interaction.tex
One of the key areas in autonomous robots is teaching robots to interact with the environment. A robot may be equipped with one or several end-effectors, such as parallel jaw grippers, suction cups, etc., which can be used to modify the state of its surroundings by moving or pushing objects. While the principle of robotics manipulation for known object geometry is well studied~\cite{murray_mathematical_2017}, a significant effort is still required for applying it to 6-DoF manipulation of cluttered real objects which is affected by various uncertain factors such as sensor noise, mirror reflections, etc.

\subsection{Open-loop Grasping}

In the task of grasping an object, a robotic hand must understand its geometric relationship with the target, specifically, the distance to the object and its pose in six degrees of freedom (6-DoF), which includes three axes of translation and three axes of rotation. Traditionally, this information is obtained using range sensors, such as Lidar or depth cameras, or it can be estimated from RGB images using pose estimation methods. Once observed, this data must be converted into an explicit 3d representation, such as point clouds or meshes. This step is crucial for generating plausible grasps, as well as for planning the robot's trajectory to execute the task.

Grasping can be decomposed into three stages, an observation stage where the robot perceives the target objects, a planning stage where the robot generates grasp proposals and an execution stage where the grasping is carried out by the robot. This ``perceive-plan-execute'' strategy works fairly well for static objects.

NeRF can be a powerful tool to model the 3D world for robotic interaction. The most straightforward solution is a 'loosely-coupled' NeRF integration into the grasping pipeline. In this setting, the NeRF model is used to create 3D geometries from multi-view images. The property of NeRF to model view-dependent visual effects makes it especially suitable for representing transparent surfaces. Some works researched grasping such objects~\cite{ichnowski_dex-nerf_2021, dai_graspnerf_2023, kerr_evo-nerf_2022, yen-chen_mira_2022}, they create a NeRF model to extract geometry and better depth maps from RGB inputs. Some use normal cues as opposed to RGB to account for view-dependent effects~\cite{lee_nfl_2024}. Some use a poking strategy to generate 3D reconstructions for unknown objects~\cite{chen_perceiving_2023}. Some use affordance information to generate the next best views for online NeRF training to maximize success rate~\cite{zhang_affordance-driven_2023}. Another way to incorporate NeRF is to use novel view synthesis and density field from NeRF to directly train multi-view descriptors for explicit grasping execution~\cite{yen-chen_nerf-supervision_2022}.

The grasping problem becomes more difficult when dealing with deformable or fragile objects, i.e., considering closed-loop feedback. The force exerted by a robot's hand needs to be controlled precisely to properly handle these targets. NeRF can be used to collect dense synthetic tactile data~\cite{zhong_touching_2022} from a few input samples for efficient neural model training.

\subsection{Closed-loop Control}

Although open-loop methods have proven effective in manipulating complex objects within cluttered scenes, they struggle to handle dynamic objects and require precise camera calibration. To overcome these limitations, several studies have focused on learning policy functions that enable more generalized environmental manipulation. In this domain, two primary methodologies have emerged: imitation learning and reinforcement learning (RL).

Imitation learning, also known as behavior cloning, distills knowledge from expert demonstrations and learns a policy function for specific tasks. The input is usually encoded by a neural network to generate smaller latent vectors, then these learned low-dimensional representations are used to directly regress candidate grasping poses. Owing to the superior rendering quality of NeRF, some work uses it to generate visual-action datasets for behavior cloning~\cite{zhou_nerf_2023}. Imitation learning is considered an offline approach since it is trained on a large amount of input-action samples.

Another way to achieve closed-loop control of robot manipulation is through deep reinforcement learning. In this setting, the environment and robot state is modeled as a Markov Decision Process. Given a policy and a reward function, the robot learns to achieve its goal by interacting with the environment through trial and error. Such a network is often trained using computer simulations first, then transferred to the real world. NeRF is good at generating photo-realistic synthetic views given a few input images and, therefore is suited for sim-to-real transfer learning. For example, Byravan et al.~\cite{byravan_nerf2real_2023} create a NeRF of real indoor scenes, then generate realistic novel views for policy learning from simulations. 

A less obvious approach is to use NeRF to distill high-dimensional data into low-dimensional latent vectors, as RL often cannot directly operate on raw input data. Works such as~\cite{driess_reinforcement_2022} train a NeRF network as a feature encoder to encode view-independent features. Then they freeze the NeRF network and train a policy network to decode such features into actions. A similar method appears in~\cite{shim_snerl_2023} but with semantic cues.

Overall, NeRF has shown promising performance in robotic manipulation but remains under-explored. One interesting line of new research is to make grasping selections more sensible for humans using learned language models~\cite{rashid_language_2023}.

%% file: sections/11_future_works.tex
\subsection{Gaussian Splatting}
\label{subsec::gs}

Recently proposed 3DGS method has revolutionized radiance fields with its very fast and high-quality renderings. Gaussian map $\mathcal{G}=(\mathbf{p}, \mathbf{c}, o, \mathbf{\Sigma})$ is represented by a set of unordered points, each with their attributes such as position $\mathbf{p}$, color $\mathbf{c}$, opacity $o$ and covariance $\mathbf{\Sigma}$. The 3D Gaussian shapes can be represented as $G(x)=\exp^{-\frac{1}{2}x^T\Sigma^{-1}x}$. To render Gaussians, points are first projected to the screen space with the affine approximation $\Sigma^{\prime}=JW\Sigma W^TJ^T$, where $J$ is the Jacobian of the projective transformation and $W$ the world transform matrix. Then a differentiable rasterization method is used to blend Gaussians that fall into the same pixel based on their opacity and distance to the screen.

Compared with NeRF that can take hours to train, 3DGS usually convergences within several minutes~\cite{kerbl_3d_2023}, which opens its door to real-time autonomous robot applications. The number of publications has surged in the past year regarding various of autonomous robots aspects, such as novel view synthesis~\cite{kerbl_3d_2023, zheng_gps-gaussian_2023, charatan_pixelsplat_2023, lee_compact_2024}
, pose estimation~\cite{sun_icomma_2023}, feature tracking~\cite{luiten_dynamic_2023}, SLAM~\cite{matsuki_gaussian_2024, hhuang2024photoslam,yan2024gs, keetha2024splatam}
, 3D reconstruction~\cite{guedon_2023_sugar, xu_agg_2024}, robotic interaction~\cite{pokhariya_manus_2023}, autonomous driving~\cite{zhou_drivinggaussian_2023}, etc.

Specifically speaking, 3DGS can be used in the following ways to improve upon existing NeRF methods:

\begin{itemize}
    \item Enhanced novel view synthesis capability: Although vanilla NeRF and subsequent variants revolutionized NVS with the concept of rendering radiance fields, its novel view synthesis capability is still limited compared to what can be achieved with 3DGS~\cite{kerbl_3d_2023}. This is especially true for online applications such as SLAM, current works of GS-SLAM can easily surpass NeRF-based solutions~\cite{keetha2024splatam}, and with even faster operational speed.
    \item Robust scene reconstruction from monocular images: Reconstructing a scene from multi-view images has long been a difficult task due to the ill-posedness of the problem. 3DGS proves that even with a handful of randomly distributed points, a robust initialization can still be achieved, removing the need to explicitly initialize the map from triangulating points.
    \item Improved convergence property for pose estimation: 3DGS is said to have a better convergence basin compared with NeRF methods~\cite{yugay_gaussian-slam_2023}, meaning the camera pose can be solved with larger deviations from the ground truth, compared with NeRF-based methods. This is beneficial for robust SLAM operation. 
\end{itemize}

However, 3DGS methods are still weak at reconstructing geometry from RGB images, since the model only cares about rendering quality. Although there are several works trying to achieve high-quality 3D reconstruction, they either try to solve this problem using explicit meshing~\cite{guedon_2023_sugar} or by regularizing Gaussian shapes and depth maps~\cite{huang20242dgaussian}, a coherent 3D reconstruction method from unordered Gaussian points is yet discovered.

Another problem that prevents 3DGS from applying to real-time robotic applications is the long training time. Although 3DGS has achieved much shorter training time than most NeRF methods, the time it takes to learn a moderate-sized scene representation is still challenging for online robots. For example, in~\cite{yugay_gaussian-slam_2023} they measure a $2\sim3hz$ tracking time, which is far slower than that of a typical camera input frequency. An interesting future work would be to explore acceleration techniques for real-time mapping.



\subsection{Large-Language Models}
\label{subsec::llm}

LLMs also have the potential to improve scene understanding of NeRF. One of the long-term goals of autonomous robotics is to interact with robots using natural languages. Not only does it require robots to understand human languages, but also how to use the language to describe and encode scene elements. Previous studies mostly focus on training a network on text-object mappings with a set of predefined object priors, where the number of objects that can be recognized and manipulated is limited to those in the training dataset. Generalization of out-of-distribution data is difficult and can result in poor performance. 


The recently published foundation model CLIP~\cite{radford2021learning} has proved that language models trained on a collection of captioned images can learn the relationship between textual and visual elements, which can in turn be used to link texts with visual scenes. This concept can be naturally extended to NeRF with a few modifications. LERF~\cite{kerr23lerf} distills CLIP features into NeRF, making textual queries of the NeRF scene possible. LERF-TOGO~\cite{rashid:2023:lerf2go} further improves LERF with DINO~\cite{caron2021emerging} for zero-shot semantic grasping. OV-NeRF~\cite{ovnerf} leverages the power of SAM~\cite{kirillov2023segany} to enhance open-set semantic scene understanding. Overall, the topic of Language-embedded NeRF is still an under-explored topic and may need more attention in the future. 


\subsection{Generative AI}
\label{subsec::gai}

NeRF is inherently a fitting method, where the model parameters are estimated from a set of input images to best explain the 3D scene. However, since every scene is reconstructed from scratch, without general knowledge of 3D, it can be difficult for NeRF to reconstruct scenes with only a few observations, which is a common situation for robot operations. The ability to generalize to unseen data can be a key enabling factor for robust scene mapping. 

Methods such as pi-GAN~\cite{chan_pi-gan_2021} and GRAF~\cite{schwarz_graf_2021} pioneered the use of NeRF with GAN for 3D aware content generation, but they are limited to generating 3D human portraits. VAE-based methods try to improve the reconstruction quality and training stability using variational auto-encoders~\cite{kosiorek2021nerfvae}, while can generate out-of-distribution data, while GAN-based methods cannot.

Recently, diffusion models are making great progress toward photorealistically generating images from text prompts. Therefore, one interesting line of research is trying to bring the power of denoising diffusion models to NeRF applications. For example, DiffRF~\cite{yu_edit-diffnerf_2023} proposes to apply diffusion on radiance fields, obtaining much higher quality reconstruction than GANs. PoseDiff~\cite{seoyoung_lee_posediff_nodate} uses diffusion and text conditioning models to complete scene contents with only a sparse set of images. Shap-E~\cite{jun_shap-e_2023} uses a diffusion model to generate multi-modal implicit fields with both texture and geometry features. Another interesting research direction is to use diffusion models for pose estimation~\cite{cheng_id-pose_2023}.

The biggest problem when combining NeRF and generative AI is the limited computational resources. While NeRF might be widely regarded as a memory-efficient 3D scene representation, the memory and computational power needed for training such NeRF models are still prohibitive. This situation is getting worse when generative AI can also consume huge resources, therefore this method is infeasible for large scenes. Speeding-up training and scaling-up to larger scenes are explored by subsequent works, using techniques discovered in recent years for efficient NeRF reconstruction, such as sparse voxels~\cite{schwarz2022voxgraf} and tri-planes~\cite{chan2022efficient}, but reconstruction of large environments is yet possible for these methods.

%% file: sections/12_conclusion.tex
\label{sec:conclusion}
This survey provides a comprehensive overview of NeRF and their applications in autonomous robotics, emphasizing how they can address various challenges in robotic perception, localization and navigation, decision-making. NeRF has been highlighted as an approach for high-fidelity 3D scene modeling, its capability to produce detailed reconstructions from sparse data allows autonomous robots to better understand and navigate complex environments. The discussion extends to various enhancements of NeRF, including handling dynamic scenes, estimating poses and integrating multi-modal sensory data for localization and mapping, which are pivotal for improving the accuracy and adaptability of robotic systems.

Despite the advantages, NeRF still faces challenges such as computational demands and the requirement for substantial training data. The survey proposes future research directions aimed at overcoming these obstacles to fully leverage NeRF’s potential in robotics. As the technology matures, NeRF is expected to significantly advance autonomous systems, promising breakthroughs in autonomous robotic perception and interaction that could redefine current capabilities.